\documentclass[conference]{IEEEtran}

\usepackage[available,functional,reproduced]{ndssbadges}

\usepackage{cite}
\usepackage{amsmath,amssymb,amsfonts}
\usepackage{url}
\usepackage[numbers]{natbib}

\usepackage{paralist}

\usepackage{graphicx}
\usepackage{xspace}
\usepackage[ruled,linesnumbered]{algorithm2e}
\usepackage{xcolor}
\usepackage{multirow}
\usepackage{booktabs}
\usepackage{subfigure}
\usepackage[para,online,flushleft]{threeparttable}
\usepackage{hyphenat}
\usepackage[backref=page]{hyperref}
\usepackage{pifont}
\newcommand{\cmark}{\ding{51}}%

\graphicspath{{./figures/}}

\newcommand{\ours}{\texttt{CamPro}\xspace}

\begin{document}

\title{\huge \ours: Camera-based Anti-Facial Recognition}

\author{
    \IEEEauthorblockN{Wenjun Zhu, Yuan Sun, Jiani Liu, Yushi Cheng, Xiaoyu Ji*, Wenyuan Xu}
    \IEEEauthorblockA{USSLAB, Zhejiang University\\ 
    \{zwj\_,sy\_tsang,jianiliu,yushicheng,xji,wyxu\}@zju.edu.cn}
}

\IEEEoverridecommandlockouts
\makeatletter\def\@IEEEpubidpullup{6.5\baselineskip}\makeatother
\IEEEpubid{\parbox{\columnwidth}{
    Network and Distributed System Security (NDSS) Symposium 2024\\
    26 February - 1 March 2024, San Diego, CA, USA\\
    ISBN 1-891562-93-2\\
    https://dx.doi.org/10.14722/ndss.2024.24158\\
    www.ndss-symposium.org
}
\hspace{\columnsep}\makebox[\columnwidth]{}}

\maketitle

\footnotetext[1]{Xiaoyu Ji is the corresponding author.}
\footnotetext[2]{Artifact: \url{https://doi.org/10.5281/zenodo.10156141}}
\footnotetext[3]{Code Release: \url{https://github.com/forget2save/CamPro}}


\begin{abstract}

The proliferation of images captured from millions of cameras and the advancement of facial recognition (FR) technology have made the abuse of FR a severe privacy threat. Existing works typically rely on obfuscation, synthesis, or adversarial examples to modify faces in images to achieve anti-facial recognition (AFR). However, the unmodified images captured by camera modules that contain sensitive personally identifiable information (PII) could still be leaked. In this paper, we propose a novel approach, \ours, to capture inborn AFR images. \ours enables well-packed commodity camera modules to produce images that contain little PII and yet still contain enough information to support other non-sensitive vision applications, such as person detection. Specifically, \ours tunes the configuration setup inside the camera image signal processor (ISP), i.e., color correction matrix and gamma correction, to achieve AFR, and designs an image enhancer to keep the image quality for possible human viewers. We implemented and validated \ours on a proof-of-concept camera, and our experiments demonstrate its effectiveness on ten state-of-the-art black-box FR models. The results show that \ours images can significantly reduce face identification accuracy to 0.3\% while having little impact on the targeted non-sensitive vision application. Furthermore, we find that \ours is resilient to adaptive attackers who have re-trained their FR models using images generated by \ours, even with full knowledge of privacy-preserving ISP parameters.





\end{abstract}

\section{Introduction}

The rapid development of DNN has facilitated various computer vision applications that recognize human activity, such as person detection~\cite{yolov5}, human pose estimation~\cite{mmpose2020}, and image caption~\cite{rennie2017self}, in areas such as surveillance~\cite{nikouei2018smart}, healthcare~\cite{Chen2021FallDS}, sports~\cite{Wang2019AICD}, fitness~\cite{aifitness}, etc. 
However, the sensitive personally identifiable information (PII), especially the faces in the images~\cite{pii}, is simultaneously collected and uploaded to untrusted third-party servers. The recent advance in facial recognition (FR) techniques~\cite{schroff2015facenet,deng2019arcface,Meng2021MagFaceAU,Kim2022AdaFaceQA} has made it easy and cheap to identify people by their faces. As reported by the National Institute of Standards and Technology (NIST), the face identification accuracy on common webcam images is up to 99.35\% among 1.6 million people~\cite{frvt}. That has made the abuse of FR or even stalking~\cite{shwayder2020clearview} possible, resulting in numerous lawsuits~\cite{patrick2021historic, joe2021tiktok, shwayder2020clearview}. The fear of privacy infringements has caused reluctance to adopt CCTV in European countries for years~\cite{Calipsa}, and both California and Portland have even banned FR techniques~\cite{pardau2018california, hatmaker2020portland}. 
Nevertheless, this paper aims to enable people to benefit from modern technologies, i.e., freely using vision applications, while preserving their privacy, i.e., achieving the goal of anti-facial recognition (AFR)~\cite{wenger2022sok}.

\begin{figure}
    \centering
    \includegraphics[width=.98\linewidth]{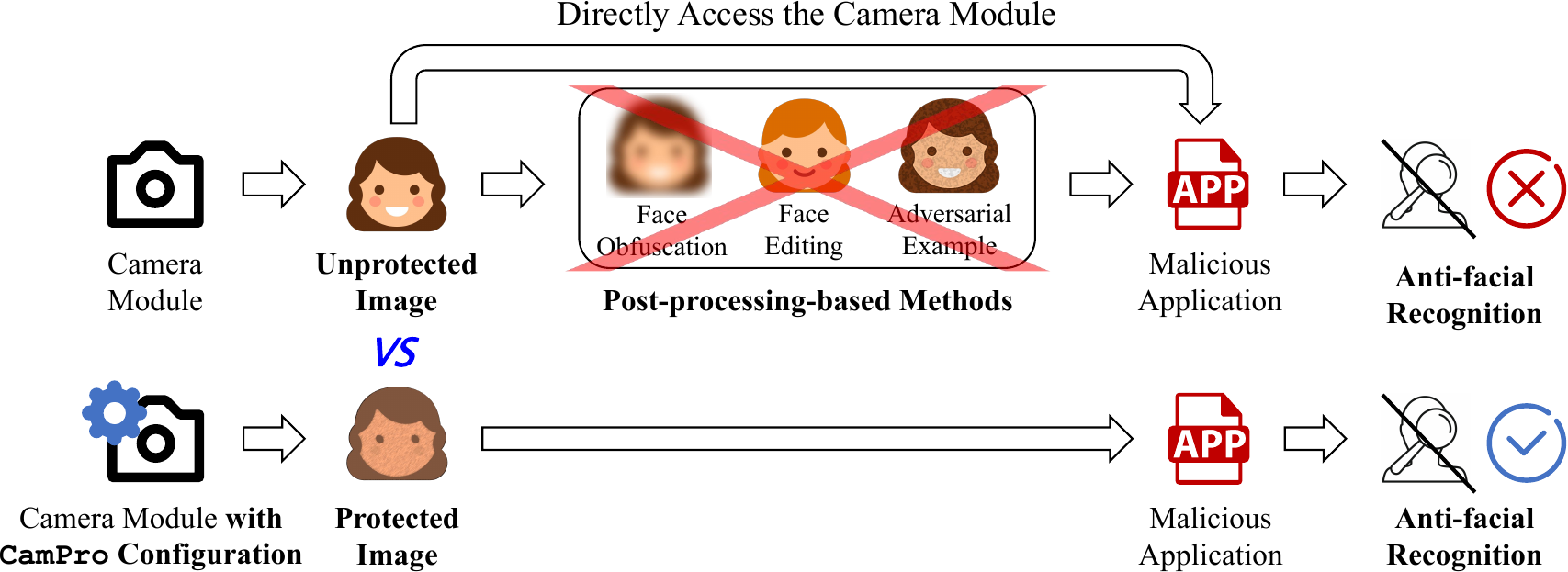}
    \caption{\ours resides inside a camera module to achieve anti-facial recognition (AFR) during the generation of images, i.e., \emph{privacy-preserving by birth}, while traditional AFR methods desensitize the raw images output by the camera module, i.e., based on post-processing.}
    \label{fig:intro}
\end{figure}


Existing literature works on the output images of the camera module to achieve AFR, i.e., relying on post-processing. Representative solutions include face obfuscation~\cite{Yang2022ASO, Ilia2015FaceOffPP, Koshimizu2006FactorsOT}, face editing~\cite{Bitouk2008FaceSA, Rhee2013CartoonlikeAG, Kuang2021EffectiveDG}, and facial adversarial examples~\cite{shan2020fawkes, cherepanova2021lowkey, Yang2021TowardsFE, yin2021adv}, etc. 
For those post-processing-based solutions, an adversary can bypass the protection if she can directly access the camera module to obtain raw images, as depicted in Fig.~\ref{fig:intro}. 
The adversary may access the camera module in two ways, i.e., (1) allowed as requested permission of the normal operation, and (2) illegally achieved by compromising the operating system (OS).
This inspires us to ask one research question, ``\emph{Is it possible to achieve AFR inside the camera module, which is isolated from the OS and hopefully can be difficultly compromised?}''

To this end, we propose the concept of \emph{privacy-preserving by birth}, i.e., camera modules shall generate protected images with PII removed to achieve AFR, which yet shall contain enough information to support the targeted vision application, such as person detection. 
By transferring the process of privacy protection from outside the camera module to inside it, the attacker's opportunity to bypass the protection is limited as the image acquisition and privacy protection are bound together.
We design and implement \ours, which achieves AFR inside a camera module but without the need of modifying the hardware of existing commodity camera modules.


The goal of \ours is challenging in terms of how to manipulate the image acquisition to achieve the balance between privacy protection and utility preservation of images.
Firstly, the original design purpose of the camera module is to capture images that are consistent with human perception; hence, there are no existing privacy-preserving functions inside the camera module. Although there are built-in image signal processing (ISP) functions, e.g., demosaicing, gamma correction (Gamma), color correction matrix (CCM), etc., it is unclear whether any of them can provide the capability of AFR.
Secondly, even if we can find proper ISP functions, it is necessary to guarantee that the modification by \ours should have little effects upon the normal operation of the targeted vision application, e.g., person detection. Since the ISP functions are mostly image-agnostic and globally applied, it is not trivial to find appropriate ISP parameters that prevent facial recognition but allow the targeted vision application.
Thirdly, the appearance of \ours images should also be taken into consideration, especially for a few usages, such as surveillance, where the images for visual recognition may be viewed by humans for a second check or other purposes.


To address above challenges, we first conducted extensive analysis and found that color-related ISP functions, such as Gamma and CCM, are good candidates for achieving AFR due to their non-linear mapping mechanism at the pixel level. Then, to strike a balance between privacy preservation and the utility of vision applications, we design an adversarial learning framework to find appropriate parameters for those ISP functions. Last but not least, to provide useful visual information in a more human-friendly format, we design a privacy-preserving image enhancer to reconstruct the images without harming the privacy preservation.
We implemented and validated the prototype of \ours on a commercial camera module. Our experiments demonstrate that \ours can achieve a success rate of 99.7\% against the black-box state-of-the-art face identification systems, with little influence on the performance of person detection. Our contributions in this paper are summarized below:
\begin{itemize}
    \item We propose a new paradigm to preserve privacy by birth that enables common camera modules to output protected images without hardware modification.
    \item We propose to use built-in ISP functions to desensitize images, employ an adversarial learning framework to optimize the ISP parameters to satisfy the design requirements of both privacy and utility, and design an image enhancer to improve the visual appearance for human viewers.
    \item We validated the effectiveness of \ours on 10 state-of-the-art FR models, including the security analysis on white-box adaptive attacks. As a proof-of-concept, we implemented \ours on a commercial camera module and validated it in the real world.
\end{itemize}


\section{Background}

In this section, we present the background knowledge of (1) facial recognition techniques that we want to disable, (2) vision-based human activity recognition whose utility we want to maintain, and (3) the camera module that we use to achieve our goals.

\subsection{Facial Recognition Techniques}
\label{sec:identification}

\begin{figure}
    \centering
    \includegraphics[width=.98\linewidth]{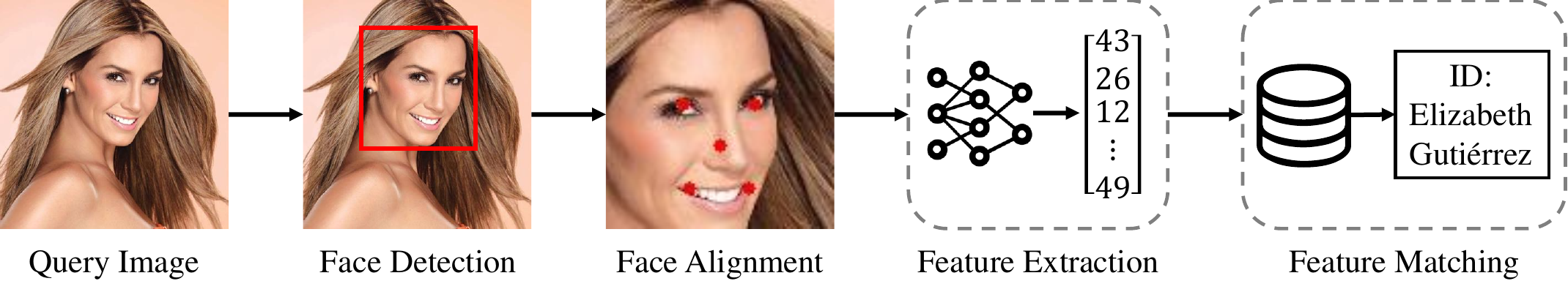}
    \caption{Typical stages of face identification.}
    \label{fig:fr}
\end{figure}

Facial recognition (FR) identifies or verifies a person's identity based on their facial features. As a long-standing topic in the field of computer vision, FR techniques have been updated for many generations~\cite{fraden2010handbook}, and most state-of-the-art FR models are based on deep learning methodology~\cite{du2022elements}. In this paper, we focus on face identification systems (FIS) rather than face verification systems due to privacy concerns~\cite{fraden2010handbook}. In the following, we briefly introduce the typical stages of face identification, as illustrated in Fig.~\ref{fig:fr}:
\begin{enumerate}
    \item \textbf{Face Detection and Alignment.} The FIS first detects and crops the face for identification in the \emph{query image}. Then, the face is transformed to a canonical pose, i.e., aligned, according to detected facial landmarks~\cite{zhang2016joint}.
    \item \textbf{Feature Extraction.} Then, the facial feature vector is extracted with a DNN, i.e., a \emph{FR model}, for face identification. If two images belong to the same identity, the distance between their feature vectors will be close and vice versa~\cite{schroff2015facenet}.
    \item \textbf{Feature Matching.} Finally, the feature vector is matching in the \emph{gallery set} that refers to the collection of labeled facial images. There are various ways of feature matching, e.g., nearest neighbor and linear classifier. In nearest neighbor, the distances, e.g., cosine distances, between feature vectors are computed and the identity in the gallery set who behaves the nearest distance is matched.
\end{enumerate}

\subsection{Vision-based Human Activity Recognition}
Vision-based human activity recognition (HAR) is to automatically interpret human motion based on the sequences of images or the video, which can enable surveillance, healthcare, sports, fitness, human-computer interface, etc. Since images are affordable and easy to collect compared to the data of wearable sensors, the vision-based approach becomes a major branch of HAR~\cite{dang2020sensor}. However, privacy concerns about the leak of sensitive personally identifiable information (PII), especially for facial images, have become one of its major drawbacks~\cite{chen2021deep}. In this paper, \textbf{we view the vision-based HAR as the targeted vision application for which we aim to find AFR solutions.} Specifically, we investigate three representative vision applications of HAR as follows:
\begin{itemize}
    \item \textbf{Person Detection} locates all the people who appeared in an image. It can count the number of people, and furthermore,  track the movements of people by recognizing several continuous video frames~\cite{punn2020monitoring}.
    \item \textbf{Human Pose Estimation}  detects and classifies the key points of the human body, e.g., shoulders, elbows, and knees, in an image. It can analyze the movements of the user while doing an exercise or detect the injury like falling for elderly adults or people with disabilities~\cite{Chen2021FallDS}.
    \item \textbf{Image Captioning} is a multi-modal task that describes the scene of an image with natural language~\cite{rennie2017self}. It can describe the activity of humans and help identify potential crimes by detecting descriptions of suspicious behaviors.
\end{itemize}

\subsection{Camera Module}

\begin{figure}[pt]
    \centering
    \includegraphics[width=\linewidth]{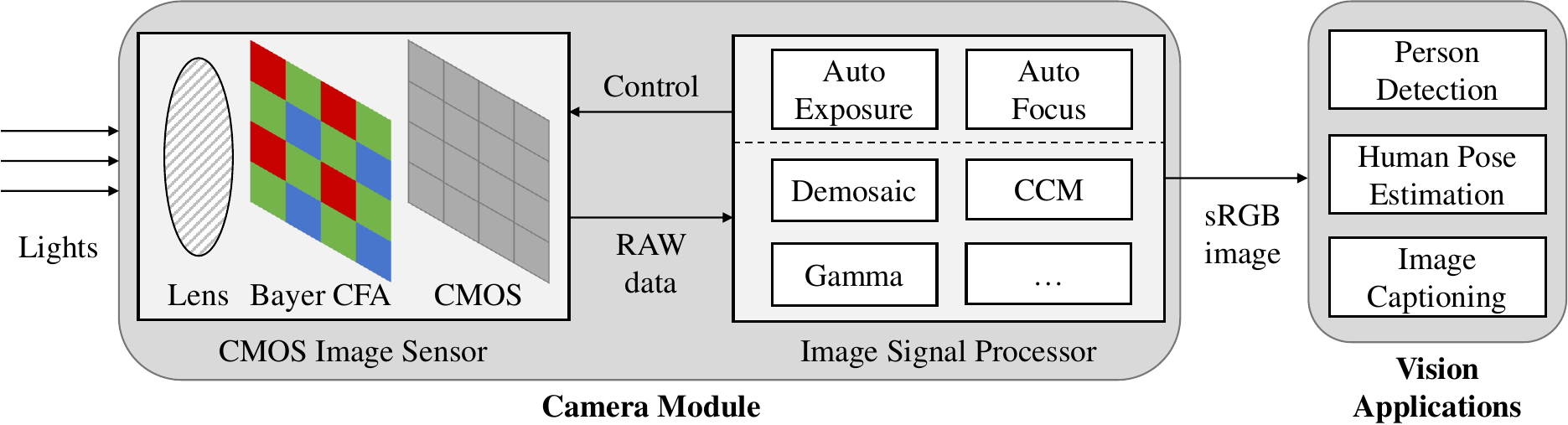}
    \caption{A camera module typically consists of an image sensor and an image signal processor (ISP). 
    \ours achieves privacy protection by tuning the parameters of ISP functions in the common camera module.}
    \label{fig:isp}
\end{figure}
A camera module usually consists of an image sensor (CMOS or CCD) and an image signal processor (ISP), as shown in Fig.~\ref{fig:isp}. 
The image sensor converts the perceived lights to raw readings (RAW), and then, the ISP, a specialized hardware for signal processing, converts the RAW to a standard RGB (sRGB) image~\cite{anderson1996proposal} that accords with human visual systems. 
The ISP is essential for modern digital cameras because (1) it provides an efficient RAW-to-sRGB conversion, and (2) it deeply involves into the control of the image sensor, for instance, adjusting the shutter and ISO to achieve the automatic exposure (AE) via a closed-loop control, as shown in Fig.~\ref{fig:isp}. 

A number of common functions are employed by most ISPs, including (1) demosaic that aggregates the channels of neighbor pixels to reconstruct a full-color image, (2) color correction matrix (CCM) that adjusts the colors with a linear transformation to be consistent with human perception, and (3) gamma correction (gamma) that is used to encode linear luminance to match the non-linear characteristics of human perception~\cite{gamma}. 
Moreover, due to the decoupled design of the image sensor and ISP, ISPs often provide a set of tunable parameters to cater to different sensors. In this paper, \textbf{we utilize the tunable parameters of ISP to enable the camera module with the capability of AFR.}

\section{Threat Model}

In this paper, we motivate to remove the sensitive personally identifiable information (PII) in a human-involved image, i.e., achieving anti-facial recognition (AFR), but maintain useful information for targeted vision applications, e.g., human activity recognition (HAR). More importantly, we aim to achieve this during the generation of images inside the camera module, i.e., \emph{preserving privacy by birth}. In the following, we present the attack model, the capability and the design requirements of our protection, named \ours.

\subsection{Attack Model}
The attacker, either individual or company, wants to identify the victim with her facial images for various malicious purposes, e.g., cybercrime, stalking, fraud, etc. The attacker mainly utilizes automatic facial recognition (FR) techniques, such as FaceNet~\cite{schroff2015facenet}, ArcFace~\cite{deng2019arcface}, etc., to perform identification since FR is scalable and even more accurate than human beings. The images that contain faces are collected by malicious apps that can directly obtain images from the camera module of the victim's device. Malicious apps may access the camera module in two ways, i.e., (1) allowed as requested permission of the normal operation, and (2) illegally achieved by compromising the OS. 

Moreover, the attacker may design \emph{adaptive attacks}~\cite{radiya2022data} against \ours when she is aware of it. Specifically, she can use existing methods, such as image restoration and model re-training, to improve the accuracy of face identification on protected images. We assume that the attacker can achieve the same type of privacy-preserving camera module as the victim's for reference, or in the worst case, know the configured ISP parameters, which provides prior knowledge of \ours for the design of adaptive attacks.



\subsection{\ours Capability and Design Requirements}

\noindent \textbf{\ours Capability.} To combat the attacker who directly obtains images from the camera module, \ours can at most utilize the built-in functions of the image signal processor (ISP) inside a common camera module to achieve AFR. Specifically, we assume that \ours can tune the parameters of those built-in functions. It is reasonable because modern ISPs have a number of tunable parameters to be compatible with various image sensors~\cite{Tseng2019Hyper}. After deployment, \ours can disable the read or write access to the ISP parameters by modifying the low-level interfaces in the firmware. Though the capability to achieve AFR is restricted within the camera module, \ours can employ any post-processing, relying on external computation resources, to maintain the utility of targeted vision application and human perception.

\noindent \textbf{Design Requirements.} Firstly, \ours shall be practical and supported by most existing commodity camera modules. Secondly, \ours shall achieve AFR during the image acquisition, indicated by significantly lowering the face identification accuracy on the output of the camera module. Thirdly, \ours shall maintain the utility of the non-sensitive targeted vision application, specifically HAR in this paper. Lastly, \ours shall make the images friendly for human viewers to perceive useful visual information, e.g., human activities, except the facial information. It is because we aim to achieve AFR against both automatic programs and human beings.


\section{Preliminary Analysis}
\label{sec:preliminary}

\begin{figure}[tbp]
    \centering
    \subfigure[raw image]{
        \includegraphics[width=.47\linewidth]{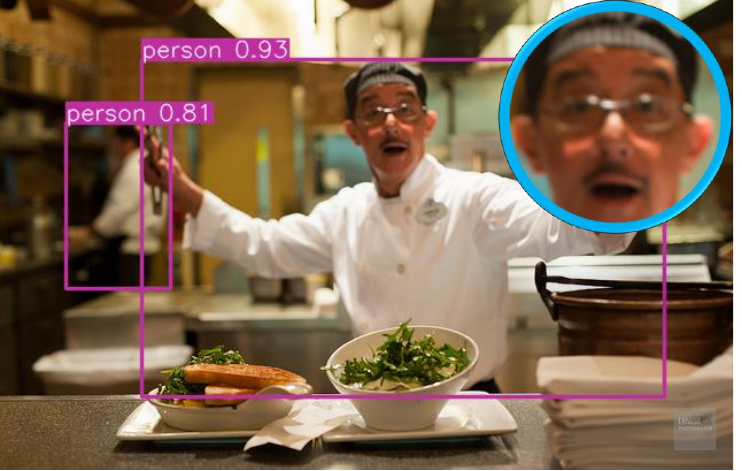}}
    \hfill
    \subfigure[color-inverted image]{
        \includegraphics[width=.47\linewidth]{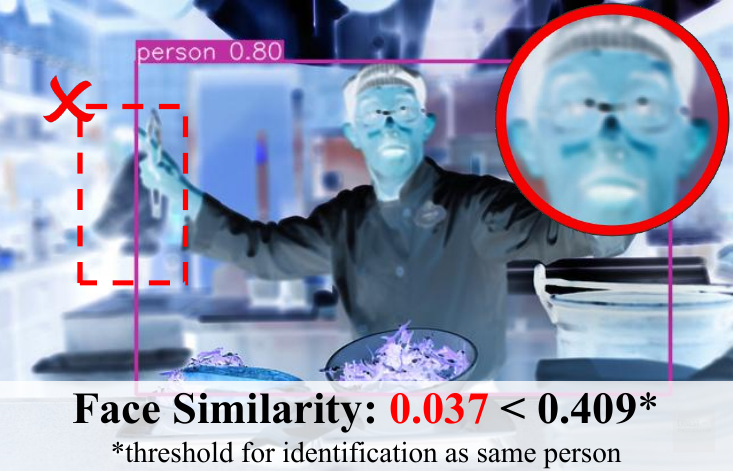}}
    \caption{Color inversion effects on FR and HAR. \emph{FR}: Faces highlighted in circles are compared by FaceNet, and they are not viewed as the same identity. \emph{HAR}: The front person is detected yet the back one is missed after color inversion. Color inversion affects the normal operation of HAR less.}
    \label{fig:inverse}
\end{figure}

In this section, we explore the feasibility of achieving AFR with some image transformations that can be implemented in the camera module. Intuitively, changes of skin colors may change the recognition result of FR models. However, changes of skin and cloth colors may also affect the HAR vision application, e.g., person detection. To investigate, we transform the randomly chosen 1,000 images with people from the COCO dataset~\cite{lin2014microsoft} by inverting the colors from $x$ to $1 - x$, where $x$ is the normalized image pixel value ranging from 0 to 1. To further investigate the effects on both FR and person detection, we conduct simulation experiments on those color-inverted images. For FR, we use MTCNN~\cite{zhang2016joint} to detect and align the faces in both the raw images and the color-inverted images, then utilize pre-trained FaceNet~\cite{schroff2015facenet} to extract the feature vectors of the faces, and finally compute the cosine similarity between them. For person detection, we use YOLOv5~\cite{yolov5} for recognition, and then simply compare the number of detected people in the raw images and the color-inverted ones.

A representative example is shown in Fig.~\ref{fig:inverse}, where two people appear in the image. For FR, the cosine similarity between the original face and the color-inverted face drops significantly from 1.0 to 0.037. For person detection, both two people are correctly detected in the raw image while after color inversion, the clear person is detected while the blurred person is missed. We also calculate the quantitative results for the tested 1,000 images. The average cosine similarity between the original face and the color-inverted face is 0.005, and only 1.6\% of them are considered to belong to the same identity with respect to the default threshold for 1-to-1 face comparing, i.e., 0.409~\cite{yin2021adv}. For person detection, $\sim$30\% of detections of people are missed after color inversion. Thus, color inversion can have a good performance of AFR while affecting the performance of person detection less.

\noindent \textbf{Remarks.} From the above experiment, it is promising that in-camera transformations realized by tuning the ISP parameters such as color inversion, are able to achieve privacy protection again FR while keeping the utility of the HAR vision application to some extent. However, it is still challenging to directly use color inversion because of: \textbf{(1) Low security.} As color inversion is completely invertible, the color-inverted images can be easily recovered by attackers with prior knowledge. \textbf{(2) Suboptimal configuration.} It is intuitive but suboptimal to use color inversion to achieve our goal. As ISP provides a number of color-related functions including CCM and Gamma, it is probable to achieve better performance via optimization on parameters of those functions. Therefore, in the design of \ours, we investigate more ISP functions mentioned above and aim to achieve privacy protection even against white-box attackers with an optimization-based approach.

\section{System Design}

\begin{figure*}
    \centering
    \includegraphics[width=.95\linewidth]{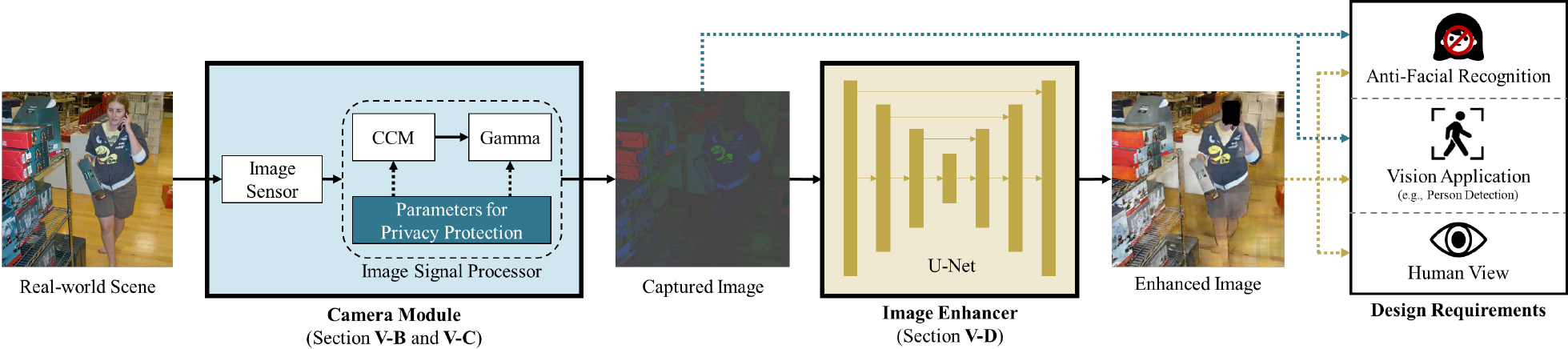}
    \caption{Overview of \ours system. \ours system consists of two main modules, i.e., (1) the camera module where the parameters of built-in ISP functions, i.e., color correction matrix (CCM) and gamma correction (Gamma), are optimized to prevent facial recognition but support the non-sensitive HAR vision application, e.g., person detection, and (2) the image enhancer that further improves the visual appearance of images for human view.}
    \label{fig:overview}
\end{figure*}

\subsection{\ours Overview}

Inspired by the preliminary analysis, we investigate the color-related ISP functions and try to find a color transformation that can eliminate the sensitive information of faces but preserve the non-sensitive information of human activities.

We design the \ours system with two main modules: (1) the \ours camera module, and (2) the \ours image enhancer, as illustrated in Fig.~\ref{fig:overview}. The outputs of the camera module and the image enhancer are denoted as \textit{captured images} and \textit{enhanced images}, respectively.
The \ours camera module refers to a commodity camera module configured with a set of optimized ISP parameters for privacy protection. A remarkable advantage of the \ours camera module is that it involves \emph{zero computational cost} since we only change the parameters of existing functions. 
The \ours image enhancer refers to a DNN that improves the visual appearance of images to provide useful information, e.g., human activities, in the captured image with a more friendly format for human view. 
According to our design, the requirements of privacy protection on AFR and utility preservation on HAR are achieved on both captured images and enhanced images, and the requirement of visual appearance is realized on enhanced images.

To fulfill the \ours system, we mainly answer the following three research questions:
\begin{enumerate}
	\item \textbf{Q1:} How to simulate the effect of the modified ISP parameters on the image such that we can optimize the parameters without actual deployments? 
    \item \textbf{Q2:} How to optimize the ISP parameters towards the goal of both privacy protection and maintaining utility, even in the face of the white-box adversary? 
    \item \textbf{Q3:} How to enhance the visual appearance of the captured image for human perception in the meanwhile preserving privacy?
\end{enumerate}
To answer \textbf{Q1},  we design the \textbf{Camera Modeling}  that models two color-related ISP functions, i.e., color correction matrix (CCM) and gamma correction (Gamma), and design a virtual imaging pipeline to simulate the effect of the modified ISP parameters on existing RGB images in the dataset. 
To resolve \textbf{Q2}, we design the \textbf{Adversarial Learning Framework} that performs a minimax optimization of face identification with both the protector and the adversary. We also add a loss component of utility to the optimization to preserve the useful information of human activities for vision applications. 
For \textbf{Q3}, we propose the \textbf{\ours Image Enhancer}, which uses a U-Net model and employs a multiple-task training scheme to obfuscate the facial region in the captured images and in the meanwhile recover other parts of the images to provide more human-friendly visual information. In the following, we introduce the three blocks in detail.



\subsection{Camera Modeling}

In this section, we simulate the effect of the modified ISP parameters on the image by camera modeling. The benefit of such an operation is two-fold: (1) It can save lots of manual efforts in capturing images with different ISP parameters in the real world. (2)  The differentiable nature of the selected ISP functions facilitates gradient-based optimization. To achieve it, we first model the selected ISP functions and then propose a virtual imaging pipeline that addresses the challenge of simulating with existing RGB images.

\subsubsection{ISP Function Modeling}

As shown in the preliminary analysis, the simple color inversion can effectively achieve AFR. To further improve the performance of AFR, we investigate all the color-related functions in common ISPs, and propose to employ the color correction matrix (CCM) and gamma correction for privacy protection.

CCM is a linear transformation for the color space conversion~\cite{finlayson2015color}, which can be  typically represented as a $3 \times 3$ matrix multiplication with 9 tunable parameters, i.e.,  $a_{11}$ to $a_{33}$:
\begin{equation}
    \begin{bmatrix}
        R_{out} \\ G_{out} \\ B_{out}
    \end{bmatrix} = \textrm{clip}_{[0, 1]}\left(
    \begin{bmatrix}
        a_{11} & a_{12} & a_{13} \\
        a_{21} & a_{22} & a_{23} \\
        a_{31} & a_{32} & a_{33}
    \end{bmatrix}
    \begin{bmatrix}
        R_{in} \\ G_{in} \\ B_{in}
    \end{bmatrix}\right)
    \label{eq:ccm}
\end{equation}
where $R_{in}$, $G_{in}$, and $B_{in}$ are the red, green, and blue pixel values of the input image, respectively. $R_{out}$, $G_{out}$, and $B_{out}$ are the corresponding pixel values of the output image after the CCM transformation. After the matrix multiplication, all the pixel values are clipped into a valid range of $[0, 1]$. 

Gamma correction is a non-linear brightness transformation that makes the output image cater to human non-linear visual systems~\cite{gamma}. Gamma correction is usually implemented as a Look-Up Table (LUT) in the ISP hardware~\cite{gammaLUT}. In commodity ISPs, the number of LUT values that can be directly configured is limited, and the other values are usually obtained by linear interpolation with those configured values. We model this mechanism via a piecewise linear function:
\begin{equation}
    y = y_i + \frac{y_{i+1}-y_i}{x_{i+1}-x_i}(x-x_i),\ i = 1,2,\cdots,k-1
    \label{eq:gamma}
\end{equation}
where $x$ denotes the input pixel value and $y$ is the output pixel value. $x_i$ and $y_i$ are the configured input values and output values, respectively. $k$ is the number of configured values. We view $x_i$ as prior constants because it depends on the interface provided by the ISP, while $y_i$ are variables, within the range of $[0, 1]$, that serve as the tunable parameters.

\subsubsection{Virtual Imaging Pipeline}
\label{sec:virtual}

\begin{figure}
    \centering
    \includegraphics[width=.95\linewidth]{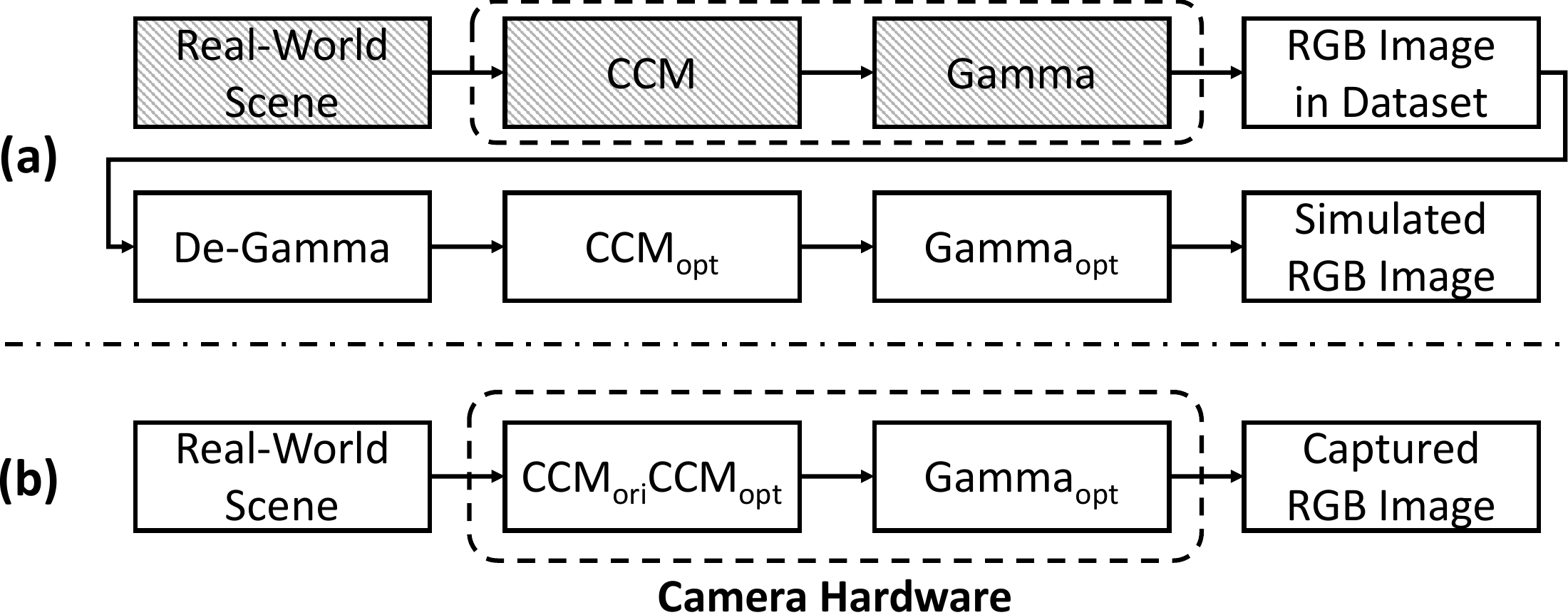}
    \caption{(a) Virtual imaging pipeline that conducts an RGB-to-RGB conversion, where shaded blocks are unknown to us. (b) Real-world imaging pipeline where the optimized ISP parameters in the simulation are deployed.}
    \label{fig:camera}
\end{figure}

With the CCM and Gamma models, we cannot directly simulate their effects on  public datasets since almost all the images in existing image datasets are RGB images that have been processed by unknown ISPs. To address it, we design a virtual imaging pipeline to approximate their effects, as shown in Fig.~\ref{fig:camera}(a). Specifically, we first apply a De-Gamma function to undo the process of Gamma, and then apply a custom CCM and Gamma function to simulate the effects of modified parameters. Note that here we use a common Gamma of 2.2 as the prior to design the De-Gamma function, i.e., $f(x) = x^{2.2}$, because we have little knowledge about the original ISP parameters for the images in datasets. We do not undo the process of CCM because there are no common parameters for CCM, which are correlated to the characteristics of the image sensor and the light condition. Instead, we reuse the original CCM of the camera module by deploying the multiplication of the original CCM and the optimized CCM, as shown in Fig.~\ref{fig:camera}(b).

\subsection{Adversarial Learning Framework}
\label{sec:advopt}

\begin{figure*}
    \centering
    \includegraphics[width=.9\linewidth]{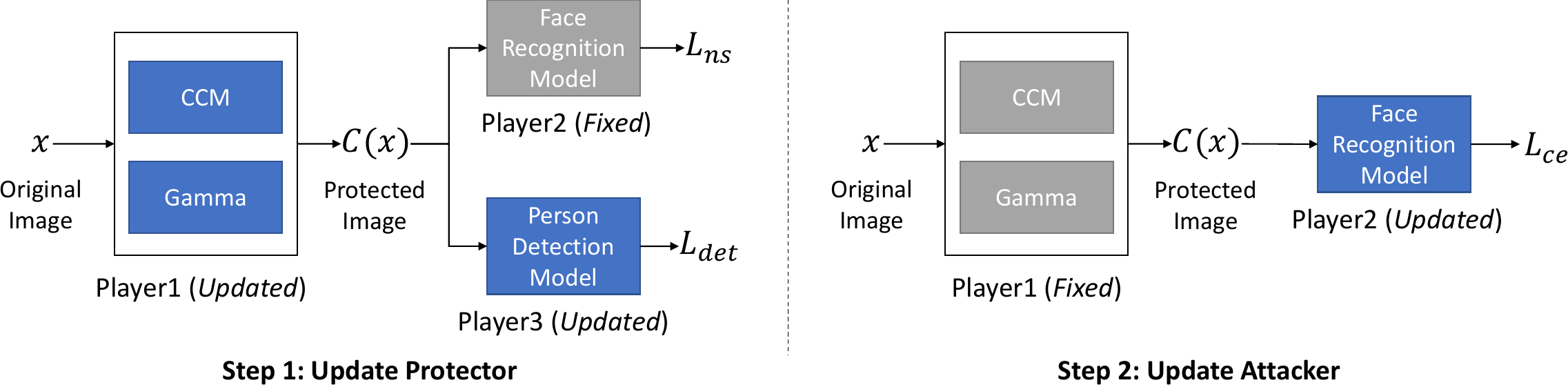}
    \caption[short]{\ours adversarial learning framework. There are three players, i.e., \textit{Player 1/2/3}, in the framework. \textit{Player1} is the camera imaging pipeline composed of CCM and Gamma. \textit{Player2} is the facial recognition model. \textit{Player3} is the targeted vision application, specifically, person detection in this paper. \textit{Player1} and \textit{Player3} are collaborators on the side of the protector while \textit{Player2} is on the side of the attacker. In \textbf{Step 1}, the parameters of \textit{Player1} and \textit{Player3} are updated to achieve a balance of privacy and utility while the ones of \textit{Player2} are fixed. In \textbf{Step 2}, the parameters of \textit{Player2} are updated to identify the protected images while the ones of \textit{Player1} are fixed. If \textbf{Step 2} is excluded, the training scheme is degraded into a similar scheme of generating adversarial examples.}
    \label{fig:advopt}
\end{figure*}

To optimize the ISP parameters with the goal of preserving privacy and maintaining utility, we propose an adversarial learning framework, where the optimization is viewed as a game of three players: (1) an imaging pipeline, controlled by CCM and Gamma parameters, i.e., the objects to be optimized,  (2) a FR model that intends to identify the faces, and (3) a HAR vision application model that indicates the utility, specifically, a person detection model in this paper. During the adversarial learning, the three players are optimized alternatively with their own loss functions until a balance of privacy and utility is achieved, as shown in Fig.~\ref{fig:advopt}. In the following, we present the adversarial learning framework in detail.

\subsubsection{Adversarial Learning Objectives}

To optimize ISP parameters, a naive approach is to employ a similar scheme of generating adversarial examples~\cite{carlini2017towards}, which enlarges the loss of the FR model by modifying the ISP parameters. However, the naive approach may give a false sense of security in the context of transferability to various FR models and robustness against adaptive attackers~\cite{radiya2022data}. We conduct an ablation study on the naive approach in Sec.~\ref{sec:ablation} and the results confirm its weak transferability over various FR models.
Here, we hypothesize that if the optimized ISP parameters can achieve privacy against an adaptive model, the privacy protection will be likely to transfer to other non-adaptive models. Therefore, we aim to achieve AFR against an adaptive attacker, i.e., a FR model fine-tuned on the \ours captured images, during the optimization of ISP parameters.
The goal of the FR model is to maximize the performance of face identification while the goal of the \ours imaging pipeline is to minimize it. Thus, they form a non-convex minimax optimization problem:
\begin{equation}
    \min_C \max_F \mathbb{E}_{(x,y) \sim \mathcal{D}_F} V(F(C(x)),y)
    \label{eq:advopt}
\end{equation}
where $C$ is the camera imaging pipeline controlled by the tunable ISP parameters and $F$ is the FR model. $\mathbb{E}$ represents the expectation value. $x$ and $y$ denote the image and label sampled from the face dataset $\mathcal{D}_F$. $V$ is a metric of face identification, e.g., accuracy.

Unlike common image classifiers, face identification systems (FIS) do not have specific classes, because every subject in the gallery set is an individual class. The composition and the size of the gallery set are variable. However, as the protector, we cannot know the gallery set of the adversary. To fulfill the optimization in Eq.~\ref{eq:advopt}, we set up a proxy classification head for the FR model, where a fixed group of people are classified. We hold the hypothesis that the effectiveness of AFR for those proxy identities would be valid for other identities in the gallery set of the adversary, as long as they are sampled from similar distributions. Thus, we convert a face identification problem into a multi-classification problem and then substitute the indifferentiable metric $V$ with the cross-entropy (CE) loss $L_{ce}$ in Eq.~\ref{eq:advopt}:
\begin{equation}
    \begin{split}
        \max_C \min_F \mathbb{E}_{(x,y) \sim \mathcal{D}_F} L_{ce} = -\log p_y \\
        p = \textrm{Softmax}(H(F(C(x))))
    \end{split}
    \label{eq:advopt2}
\end{equation}
where $p$ is a vector of probabilities and $p_y$ denotes the probability of the class $y$. $H$ is the proxy classification head that takes the features extracted by the FR model $F$ as input and outputs the logits vector.

For the optimization problem in Eq.~\ref{eq:advopt2}, a global optimum exists when $C$ converts any pixel values in the raw image into the same one, rendering that any $F$ cannot discriminate different identities. However, the solution is unacceptable because it also disables the non-sensitive vision application, e.g., person detection. Therefore, we propose to optimize for the utility simultaneously:
\begin{equation}
    \min_{C,P} \mathbb{E}_{(x,y) \sim \mathcal{D}_P} L_{det}(P(C(x)), y) = L_{cls} + L_{box}
    \label{eq:advopt3}
\end{equation}
where $P$ is the person detector. $\mathcal{D}_P$ is the dataset for person detection. $L_{det}$ represents the detection loss, composed of the classification loss $L_{cls}$ and the box localization loss $L_{box}$~\cite{redmon2018yolov3}. $L_{cls}$ indicates the difference between the predicted class probabilities and the ground-truth class labels in the image. $L_{box}$ indicates the difference between the predicted bounding box coordinates and the ground-truth bounding box coordinates.

\subsubsection{Training Scheme}

As demonstrated in Eq.~\ref{eq:advopt2}, the imaging pipeline $C$ and the FR model $F$ have opposite objectives. As a  result, a joint optimization is not feasible. Instead, we adopt alternative optimization, a common practice in generative adversarial networks (GANs)~\cite{Goodfellow2014GenerativeAN}, to solve the non-convex minimax problem. Since it is computationally unaffordable and easy to overfit for a full inner optimization, we alternate between $m$ steps of optimizing $F$ and $n$ steps of optimizing $C$. Furthermore, as formulated in Eq.~\ref{eq:advopt3}, the imaging pipeline $C$ and the person detector $P$ share the same objective on utility, so we conduct a joint optimization of them. We solve the optimization problems formulated in Eq.~\ref{eq:advopt2} and Eq.~\ref{eq:advopt3} via gradient descent. The details of the training scheme is presented in Algorithm~\ref{alg:main} in Appendix. The updated expressions of imaging pipeline, FR model, and person detection model are as follows:
\begin{equation}
    \begin{aligned}
        \theta_C & \leftarrow \theta_C - \alpha_C \nabla_{\theta_C} (L_{ns}(\theta_C, \theta_F) + \omega L_{det}(\theta_C, \theta_P)) \\
        \theta_F & \leftarrow \theta_F - \alpha_F \nabla_{\theta_F} L_{ce}(\theta_C, \theta_F) \\
        \theta_P & \leftarrow \theta_P - \alpha_P \nabla_{\theta_P} L_{det} (\theta_C, \theta_P)
    \end{aligned}
    \label{eq:grad}
\end{equation}
where $\theta_C$, $\theta_F$, and $\theta_P$ represent the parameters of the imaging pipeline, FR model, and person detection model, respectively. $\alpha_C$, $\alpha_F$, and $\alpha_P$ are the corresponding learning rates. $\omega$ is a weight to balance the privacy loss and the utility loss. $L_{ns}$ is the non-saturated CE loss~\cite{Goodfellow2014GenerativeAN} that substitutes the negative CE loss $-L_{ce} = \log p_y$, where $L_{ns} = - \log (1 - p_y)$. At the beginning of optimization, since the facial images are slightly modified, the FR model often gives high confidence prediction, i.e., $p_y \approx 1$. Thus, $L_{ce}$ is saturated such that the optimization becomes slow. At the end of optimization, the FR model cannot discriminate faces well, i.e., $p_y \approx 0$. In this case, $L_{ce}$ results in a very high loss, breaking the balance with the utility loss $L_{det}$. On the contrary, the non-saturated loss $L_{ns}$ can not only provide large gradients at the beginning of optimization but also saturate itself to balance with the utility loss at the end.

\subsection{\ours Image Enhancer}

With the \ours camera module, we can already achieve AFR while maintaining the utility from the perspective of machine perception. However, in some cases, the images captured by the \ours camera may be viewed by humans for a second check or other purposes. Considering this, we try to enhance the captured images to make them maintain a basic level of visual appearance for human perception. 

To improve the visual appearance, a simple idea is to constrain the modification of ISP parameters within a range such as an $L_\infty$ norm constraint. However, we find that slight modifications are unable to achieve a good performance of AFR. Instead of adding constraints, we propose a DNN-based image enhancer that further processes the captured images to fulfill human perception. Additionally, the enhanced images can be properly recognized by other HAR vision applications, e.g., human pose estimation and image captioning, besides the targeted one, i.e., person detection, which is involved in the adversarial learning.

\begin{figure}
    \centering
    \includegraphics[width=.95\linewidth]{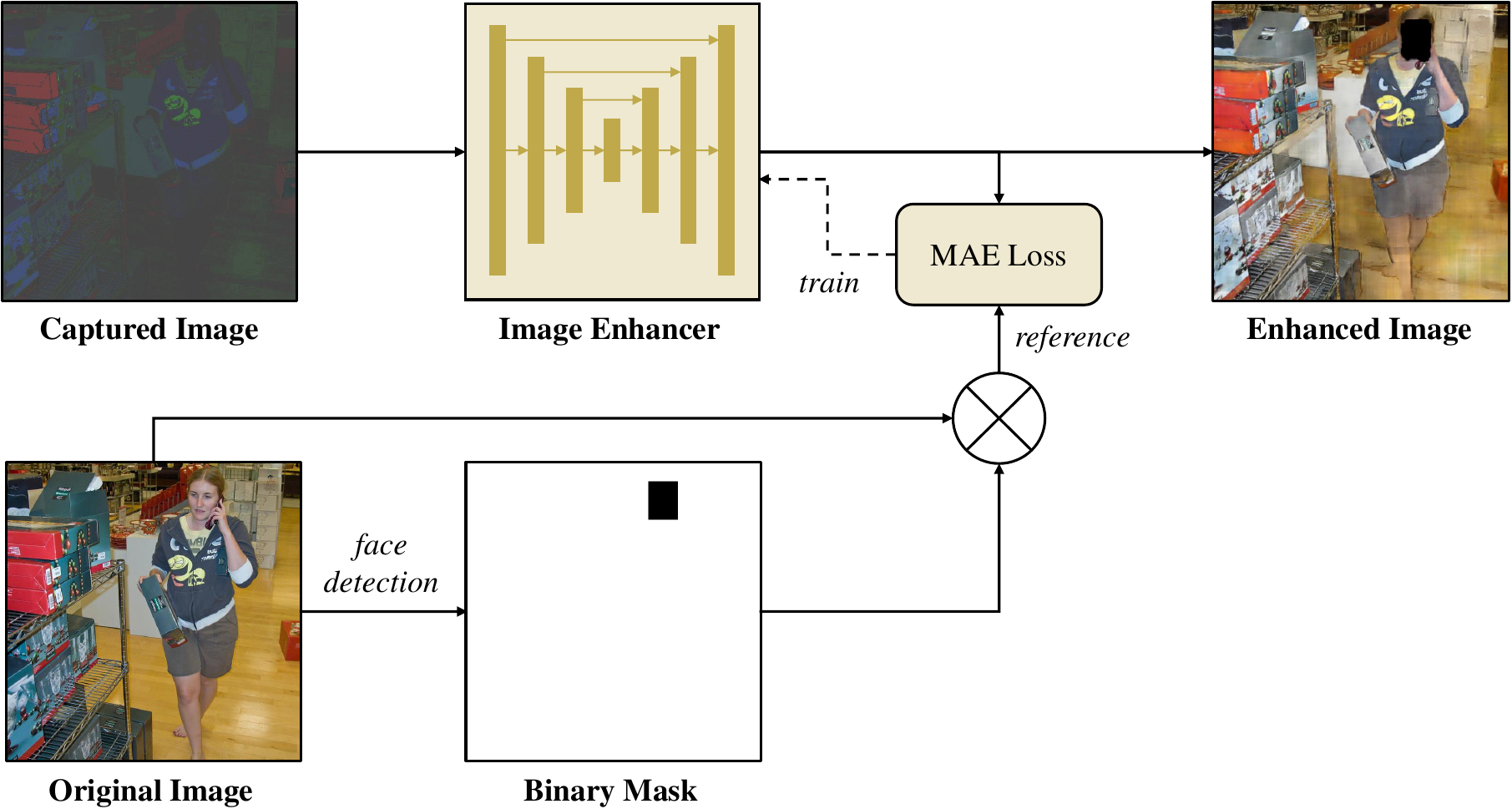}
    \caption[short]{Multiple-task training of the image enhancer. The face in the captured image is guided to be obfuscated while the other parts are reconstructed.}
    \label{fig:enhancer}
\end{figure}

To realize the image enhancer, we train a U-Net~\cite{ronneberger2015u} with pairs of original images and captured images with the goal of restoring the captured images to the original images. U-Net is a fully-convolutional network with an encoder-decoder architecture~\cite{ronneberger2015u}, which can preserve spatial information with the help of skip connections. Hence, it excels at performing a translation between images. We train the U-Net with the mean absolute error (MAE) loss, also known as $L_1$ loss. The optimization is formulated as follows:
\begin{equation}
    \min_R \mathbb{E}_{x \sim X} L_{MAE}(R(C(x)), x) = | R(C(x)) - x |
    \label{eq:restore1}
\end{equation}
where $R$ is the image enhancer and $C$ is the optimized imaging pipeline. $x$ denotes the image sampled from the standard image dataset $X$, e.g., the COCO dataset.

Since the image enhancer serves as a rough inverse transformation of the optimized imaging pipeline, the remaining sensitive information may be reconstructed. Thus, the effectiveness of \ours for privacy protection against black-box attackers may be degraded when performing FR on enhanced images. That is because the attacker gains prior knowledge from the image enhancer. To mitigate the degradation of privacy protection, we employ the multiple-task training that guides the U-Net to obfuscate the faces while reconstructing the other parts of the image. As illustrated in Fig.~\ref{fig:enhancer}, specifically, we first detect the faces in the original images, and then generate a binary mask for each original image, where the faces are filled with 0 while the others are filled with 1. Then, the MAE loss is modified as follows:
\begin{equation}
    L_{MAE} =  s \odot | R(C(x)) - x | + (1 - s) \odot | R(C(x)) |
    \label{eq:restore2}
\end{equation}
where $s$ is the binary mask of the original image $x$. $\odot$ denotes the Hadamard product operator. As shown in Fig.~\ref{fig:enhancer}, the face in the enhanced image is obfuscated while other parts of the image can be recognized by humans. Since the face obfuscation is simultaneously done with the image enhancement via a single model, it can hardly be bypassed even if the adversary can obtain the image enhancer.

\section{Evaluation}

\begin{table*}
    \centering
    \setlength{\tabcolsep}{8pt}
    \caption{Performance of Anti-facial Recognition against Face Identification Systems}
    \label{tab:privacy}
    \begin{threeparttable}
    \begin{tabular}{@{}lllrrrrrrrrrrr@{}}
        \toprule
        \multirow{2}{*}{Dataset} & \multirow{2}{*}{Image Type} & \multirow{2}{*}{Classifier} & \multicolumn{10}{c}{Facial Recognition Model (Feature Extractor)} & \multirow{2}{*}{Average} \\ \cmidrule{4-13}
         & & &  FaceNet\tnote{0} & Arc18\tnote{1} & Arc50\tnote{2} & Arc152\tnote{3} & Mag18\tnote{4} & Mag50\tnote{5} & Mag100\tnote{6} & Ada18\tnote{7} & Ada50\tnote{8} & Ada100\tnote{9} &  \\ \midrule
         \multirow{3}{*}{CelebA} & {Raw} & Nearest & 67.1\% & 77.7\% & 82.9\% & 89.5\% & 77.5\% & 90.1\% & 90.6\% & 86.6\% & 90.2\% & 90.9\% & \textbf{84.3\%} \\
         & {Captured} & Nearest & 0.0\% & 0.0\% & 0.1\% & 0.0\% & 0.0\% & 0.1\% & 0.1\% & 0.4\% & 1.2\% & 1.5\% & \textbf{0.3\%} \\
         & {Enhanced} & Nearest & 0.2\% & 0.1\% & 0.4\% & 0.4\% & 0.1\% & 0.7\% & 0.8\% & 0.8\% & 1.3\% & 1.6\% & \textbf{0.6\%} \\ \midrule
         \multirow{3}{*}{CelebA} & {Raw} & Linear & 64.7\% & 70.1\% & 69.1\% & 86.6\% & 75.5\% & 89.5\% & 90.1\% & 82.5\% & 89.1\% & 90.2\% & \textbf{80.7\%} \\
         & {Captured} & Linear & 0.0\% & 0.0\% & 0.1\% & 0.0\% & 0.0\% & 0.1\% & 0.1\% & 0.2\% & 0.6\% & 0.9\% & \textbf{0.2\%}\\
         & {Enhanced} & Linear & 0.1\% & 0.1\% & 0.2\% & 0.2\% & 0.1\% & 0.5\% & 0.5\% & 0.4\% & 0.7\% & 1.0\% & \textbf{0.4\%}\\ \midrule
        \multirow{3}{*}{LFW} & {Raw} & Nearest & 93.9\% & 92.7\% & 97.9\% & 99.2\% & 93.0\% & 99.3\% & 99.3\% & 98.7\% & 99.3\% & 99.4\% & \textbf{97.3\%} \\
         & {Captured} & Nearest & 0.1\% & 0.1\% & 0.6\% & 0.3\% & 0.1\% & 0.3\% & 0.4\% & 1.1\% & 1.7\% & 1.6\% & \textbf{0.6\%}\\
         & {Enhanced} & Nearest & 0.8\% & 0.6\% & 2.3\% & 1.4\% & 0.8\% & 2.6\% & 2.6\% & 3.3\% & 4.8\% & 5.5\% & \textbf{2.5\%}\\ \midrule
         \multirow{3}{*}{LFW} & {Raw} & Linear & 92.2\% & 92.6\% & 97.8\% & 98.7\% & 92.0\% & 99.2\% & 99.2\% & 97.6\% & 99.1\% & 99.2\% & \textbf{96.8\%} \\
         & {Captured} & Linear & 0.2\% & 0.1\% & 0.6\% & 0.3\% & 0.1\% & 0.2\% & 0.3\% & 0.7\% & 1.2\% & 1.2\% & \textbf{0.5\%}\\
         & {Enhanced} & Linear & 0.8\% & 0.7\% & 2.4\% & 1.0\% & 0.7\% & 1.9\% & 2.0\% & 2.0\% & 3.0\% & 3.7\% & \textbf{1.8\%}\\
         \bottomrule
    \end{tabular}
    \begin{tablenotes}
        \item[0] FaceNet-InceptionResNetV1; \item[1] ArcFace-IResNet18; \item[2] ArcFace-IResNetSE50; \item[3] ArcFace-IResNet152; \item[4] MagFace-IResNet18; \item[5] MagFace-IResNet50; \item[6] MagFace-IResNet100; \item[7] AdaFace-IResNet18; \item[8] AdaFace-IResNet50; \item[9] AdaFace-IResNet100.
    \end{tablenotes}
    \end{threeparttable}
\end{table*}

\subsection{Experimental Setup}
\label{sec:setup}

In this section, we present the configuration of \ours, and the datasets, models, classifiers, metrics, and the evaluation protocol of face identification.

\noindent \textbf{\ours Configuration.}
The number of configured values in Gamma is 32, where their input values are evenly spaced from 0 to 1. In the adversarial learning framework, we use Adam~\cite{Kingma2014AdamAM} with a learning rate of ${10}^{-3}$ to optimize the camera imaging pipeline, and use SGD with a learning rate of ${10}^{-1}$ and ${10}^{-4}$ to optimize the FR model and the person detection model, respectively. The weight $\omega$ in Eq.~\ref{eq:grad} that balances the privacy loss and the utility loss is set to 0.2. The adversarial learning process lasts 500 epochs. The \ours image enhancer is trained on 5,000 COCO~\cite{lin2014microsoft} images, and RetinaFace~\cite{Deng2020RetinaFaceSM} is used to generate the binary mask for those images. The optimizer of the image enhancer is AdamW~\cite{Loshchilov2017DecoupledWD} with a learning rate of $3 \times {10}^{-4}$ and a weight decay of ${10}^{-2}$, and the training process proceeds 1,000 epochs.

\noindent \textbf{Datasets.}
We employ two public facial image datasets, i.e., CelebA~\cite{liu2015faceattributes} and LFW~\cite{huang2008labeled}, to evaluate the performance of AFR, and use the COCO~\cite{lin2014microsoft} (Common Objects in Context) dataset to evaluate the performance of person detection, human pose estimation, and image captioning which are supported by COCO. In the adversarial learning framework, we use CelebA as the face dataset and use COCO as the person dataset.

\noindent \textbf{Models.}
We evaluate the AFR capability of \ours over 10 pre-trained FR models from recent studies published at top-tier conferences of computer vision, including FaceNet-InceptionResNetV1~\cite{schroff2015facenet}, ArcFace-IResNet18/SE50/152~\cite{deng2019arcface}, MagFace-IResNet18/50/100~\cite{Meng2021MagFaceAU}, and AdaFace\hyp{}IResNet18/50/100~\cite{Kim2022AdaFaceQA}. 
Moreover, we evaluate the utility of the protected images with three different HAR models, i.e., YOLOv5~\cite{yolov5} for person detection, HRNet~\cite{sun2019deep} provided by MMPose~\cite{mmpose2020} for human pose estimation, and BLIP~\cite{Li2022BLIPBL} provided by LAVIS~\cite{li2022lavis} for image captioning.

\noindent \textbf{Classifiers.}
We investigate two representative classifiers for face identification, i.e., nearest neighbor (Nearest) and linear classifier (Linear). The former is the most common classifier in the evaluation of face identification~\cite{schroff2015facenet,deng2019arcface}, and the latter is usually investigated in previous work of AFR, e.g., Fawkes~\cite{shan2020fawkes}. We use cosine distance to decide the nearest neighbor, and use softmax loss to train the linear classifier.

\noindent \textbf{Metrics.}
For the evaluation of AFR, we use \emph{face identification accuracy} as the metric, where the prediction of the face identification system (FIS) is viewed as correct if and only if it is the same as the ground truth, also known as the top-1 accuracy. Note that \textbf{a lower accuracy indicates a better protection}. For the evaluation of HAR utility, we use the standard metric of each vision application, i.e., Average Precision (AP)~\cite{lin2014microsoft} for person detection and human pose estimation, and Consensus-based Image Description Evaluation (CIDEr)~\cite{Vedantam2014CIDErCI} for image captioning. Furthermore, to quantify the utility of human perception, we utilize four full-reference image quality assessment metrics, i.e., Root Mean Square Error (RMSE), Peak Signal-to-Noise Ratio (PSNR)~\cite{hore2010image}, Similarity Structural Index Measure (SSIM)~\cite{wang2004image}, and Multi-Scale Similarity Structural Index Measure (MS-SSIM)~\cite{wang2003multiscale}.

\noindent \textbf{Evaluation Protocol of Face Identification.}
We use a closed-set protocol of face identification for evaluation~\cite{KemelmacherShlizerman2015TheMB,Gnther2017TowardOF}. Specifically, we first filter the subjects who have more than 2 images in the dataset. Thus, the total numbers of subjects are 10,126 and 1,680 for CelebA and LFW, respectively. Then, we build the gallery set owned by the adversary by randomly choosing one image of each subject. We also create a sequence of query images by randomly choosing another image of each subject to conduct face identification. To evaluate the performance of privacy protection, the query images are supposed to be protected by a certain method, e.g., \ours or other baseline methods. To reduce the variation of the result, we run the closed-set identification protocol 10 times independently and report the mean accuracy by default.

\vspace{-3pt}
\subsection{Evaluation of Privacy Protection}
\label{sec:privacy}

In this section, we evaluate the privacy protection performance of \ours against 20 face identification systems, i.e., the combination of 10 existing FR models and 2 types of common face identification classifiers. The detailed results can be found in Table~\ref{tab:privacy}. In the following, we first present the overall performance and then analyze the impacts of various factors, e.g., FR models, face identification classifiers, and face datasets, on the performance of AFR.

\subsubsection{Overall Performance}
The adversary may exploit either the images output by the \ours camera module, i.e., the captured images, or the images output by the \ours image enhancer, i.e., the enhanced images, for face identification. Therefore, we investigate the privacy protection performance of \ours on both captured images and enhanced images. We also evaluate the face identification accuracy on the raw/unprotected images as the baseline.
We find that the face identification systems with the nearest neighbor classifier achieve an average accuracy of 84.3\% on the raw images of CelebA dataset, an average accuracy of 0.3\% on the captured images, and an average accuracy of 0.6\% on the enhanced images. The results indicate that both the captured images and the enhanced images of \ours achieve good performance of privacy protection.

\subsubsection{Impact of Various Facial Recognition Models}
We employ 10 state-of-the-art DNN models that are highly diverse in training datasets, losses, and DNN architectures. Among the 10 models, only one model, Ada18 (Model 7 in Table~\ref{tab:privacy}), is seen by \ours while the others are unseen.
From the results, we can see that the face identification accuracies achieved by the seen model are 0.4\% for the captured images and 0.8\% for the enhanced images, while the highest accuracies achieved by the unseen models are 1.5\% for the captured images and 1.6\% for the enhanced images. It indicates that the privacy-preserving effects of \ours can generalize well across black-box FR models.

\subsubsection{Impact of Various Face Identification Classifiers}
Besides the common nearest neighbor classifier used for feature matching as default, we also evaluate the softmax linear classifier trained with labeled face images. The face identification systems with the linear classifier achieve an average accuracy of 80.7\% on the raw images of CelebA dataset, an average accuracy of 0.2\% on the captured images, and an average accuracy of 0.4\% on the enhanced images, which is overall slightly lower than those with the nearest neighbor classifier. The results indicate that \ours also works on the face identification systems with the linear classifier.

\subsubsection{Impact of Various Face Datasets}
We employ two standard face datasets, including the trained one, CelebA, and the untrained one, LFW, for evaluation. The subjects between the two datasets are not overlapped.
From the results, we find that \ours can protect 99.4\% of subjects from CelebA and 97.5\% of subjects from LFW.  However, for raw images, the face identification accuracy achieved on CelebA, i.e., 84.3\%, is also lower than that on LFW, i.e., 97.3\%. It is because LFW has fewer subjects and fewer pose variations, thus it is easier to identify faces on LFW than on CelebA.

\begin{figure*}
    \centering
    \includegraphics[width=.9\linewidth]{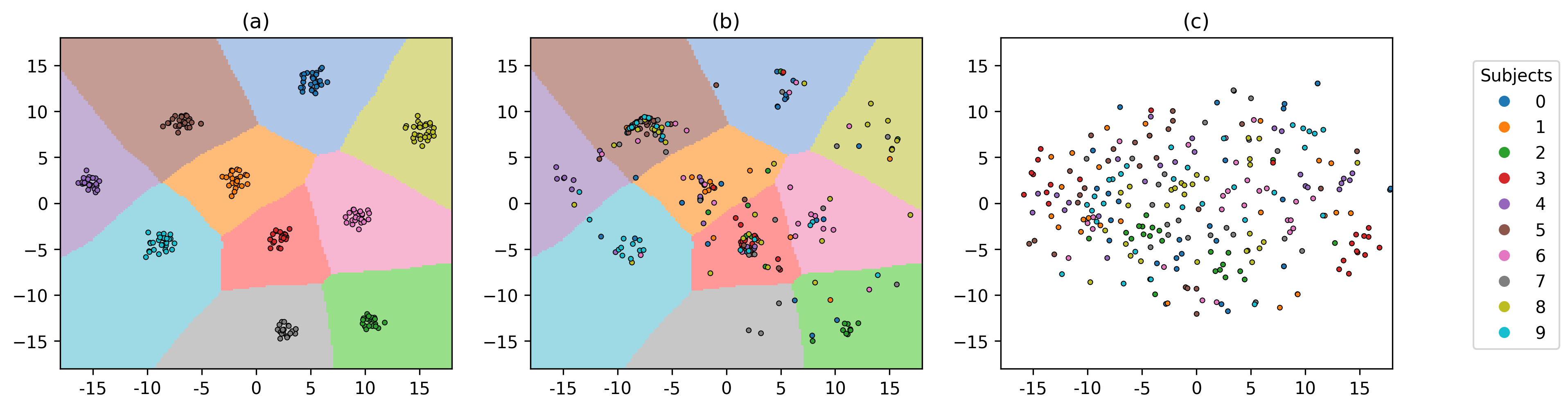}
    \caption{The t-SNE visualization of facial features. Every dot in Fig.~(a)/(b)/(c) represents an image, and its color represents the identity. In Fig.~(a) and~(b), we visualize the decision boundaries of the nearest neighbor classifier via a colorful background. Fig.~(a) shows the features of the raw images. Fig.~(b) shows the features of the protected images in the same embedding space as~(a). Fig.~(c) shows the features of the protected images in another standalone embedding space. The comparison between Fig.~(a) and~(b) demonstrates the AFR effects. The comparison between Fig.~(a) and~(c) indicates that the FR model is unable to extract good features for identification. }
    \label{fig:t-SNE}
\end{figure*}

\vspace{-2pt}
\subsection{Visualization of Facial Features}
\label{sec:feature}
To better understand the effects of \ours on the face identification systems, we use t-distributed Stochastic Neighbor Embedding (t-SNE)~\cite{van2008visualizing}, which embeds high-dimensional data into low-dimensional data (e.g., 2D) via self-supervised learning, to visualize the facial features extracted by the FR model. We randomly select 10 subjects from CelebA, and use the facial features extracted by Ada18 for visualization. Furthermore, we run the nearest neighbor algorithm with the embedded features to visualize the decision boundaries between every two different subjects.

First, we visualize the facial features of the raw images. As illustrated in Fig.~\ref{fig:t-SNE}(a), the features form 10 separated clusters, where various images of the same subject are close to each other; hence, the different subjects can be correctly matched via the nearest neighbor classifier. Then, we transform the facial features of the captured images by \ours into the same low-dimensional embeddings trained with the raw images. As shown in Fig.~\ref{fig:t-SNE}(b), the features of the captured images that belong to 10 subjects are distracted from their original clusters and move across the decision boundaries. As a result, the accuracy of the classifier, which works well on the raw images, significantly drops. Moreover, we train a t-SNE with the facial features of the captured images alone. As depicted in Fig.~\ref{fig:t-SNE}(c), the features of the 10 subjects are mixed up and do not form 10 clusters by subject as those of the raw images. The observation indicates that even if the classifier is trained on the facial features of the captured images, it is still difficult to discriminate different subjects.

\subsection{Ablation Study of Adversarial Learning Scheme}
\label{sec:ablation}
When optimizing the ISP parameters, we alternate with two steps, i.e., (1) the protector update step and (2) the attacker update step. The details of each step can be found in Sec.~\ref{sec:advopt}. The ISP parameters are not updated in the attacker update step. Therefore, we conduct an ablation study by removing the attacker update step. After removing the adversary step, the adversarial learning scheme is converted into a similar scheme with the generation of adversarial examples. We evaluate the privacy protection performance with the removal of the attacker update step, denoted as \textbf{Protector-Only}, on CelebA. We keep the other settings of optimization, e.g., only one white-box/seen model, Ada18. 

As presented in Table~\ref{tab:ablation}, we observe that both \textbf{Protector-Only} and \textbf{Protector+Attacker}, i.e., \ours, achieve good performance on the white-box model; however, \textbf{Protector-Only} can not transfer well to some black-box models, for example, Ada100 still achieves an accuracy of 40.4\%. The average accuracy of the 9 black-box models is 16.4\%, which is 1093\% of the accuracy of the white-box model, while the value is 0.3\% for \textbf{Protector+Attacker}. The ablation study suggests that the attacker update step is necessary for the adversarial learning scheme. The optimization against an adaptive model helps our protection to generalize on other black-box models.

\begin{table}[]
    \centering
\caption{Ablation Study of Adversarial Learning Scheme}
\label{tab:ablation}
\begin{threeparttable}
\begin{tabular}{@{\extracolsep{4pt}}lcrr@{}}
\toprule
 & & \textbf{Protector-Only} &  \textbf{Protector+Attacker}  \\ \midrule
Ada18\tnote{*} & White-box & \textbf{1.5\%} & \textbf{0.4\%}  \\
Arc50\tnote{*} & Black-box & 1.0\% & 0.1\%  \\
Mag18\tnote{*} & Black-box & 2.3\% & 0.0\%   \\
Arc18\tnote{*} & Black-box & 2.7\% & 0.0\%   \\
FaceNet\tnote{*} & Black-box & 4.5\% & 0.0\%   \\
Arc152\tnote{*} & Black-box & 13.1\% & 0.0\%  \\
Mag50\tnote{*} & Black-box & 23.1\% & 0.1\%  \\
Mag100\tnote{*} & Black-box & 27.1\% & 0.1\%  \\
Ada50\tnote{*} & Black-box & 33.0\% & 1.2\%   \\
Ada100\tnote{*} & Black-box & 40.4\% & 1.5\% \\ \midrule
\multicolumn{2}{l}{Average Black-box Accuracy} & \textbf{16.4\%} & \textbf{0.3\%}  \\
\multicolumn{2}{l}{Accuracy ratio (Black-box/White-box)} & \textbf{1093\%} & \textbf{75\%}  \\ \bottomrule
\end{tabular}
\begin{tablenotes}
    \item[*] The abbreviations are consistent with those shown in Table~\ref{tab:privacy}.
\end{tablenotes}
\end{threeparttable}
\end{table}


\subsection{Evaluation of Utility}

In this section, we investigate the utility maintenance of \ours from two perspectives, i.e., machine perception and human perception. In the following, we first evaluate the overall performance of person detection, i.e., the targeted vision application in this paper, on the captured images, and then assess the image quality improvements by the \ours image enhancer. Furthermore, we investigate the utility of the enhanced images generalized on other vision applications.

\subsubsection{Overall Performance of Targeted Vision Application}
\label{sec:person}
\begin{figure}
    \centering
    \subfigbottomskip=-2pt
    \subfigcapskip=-3pt
    \subfigure[raw images]{
        \begin{minipage}[t]{0.32\linewidth}
            \centering
            \includegraphics[width=\linewidth,height=0.7\textwidth]{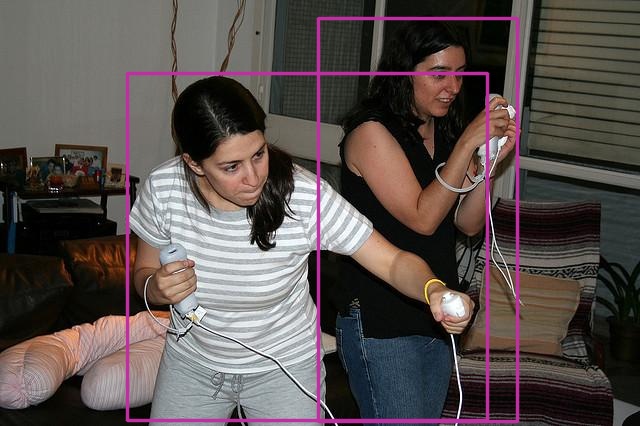}
        \end{minipage}
        \begin{minipage}[t]{0.32\linewidth}
            \centering
            \includegraphics[width=\linewidth,height=0.7\textwidth]{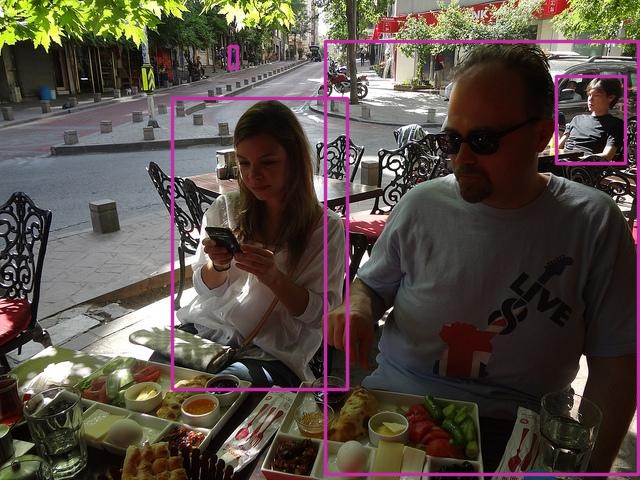}
        \end{minipage}
        \begin{minipage}[t]{0.32\linewidth}
            \centering
            \includegraphics[width=\linewidth,height=0.7\textwidth]{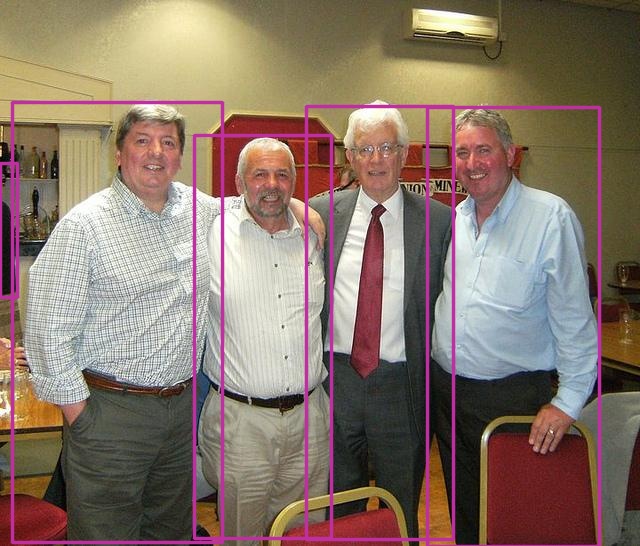}
        \end{minipage}
    }
    \subfigure[\ours captured images]{
        \begin{minipage}[t]{0.32\linewidth}
            \centering
            \includegraphics[width=\linewidth,height=0.7\textwidth]{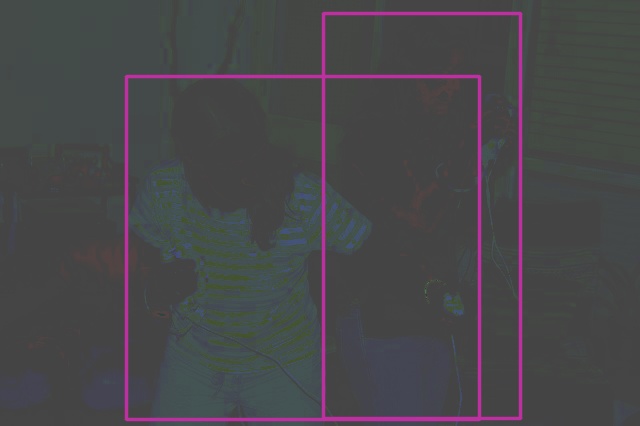}
        \end{minipage}
        \begin{minipage}[t]{0.32\linewidth}
            \centering
            \includegraphics[width=\linewidth,height=0.7\textwidth]{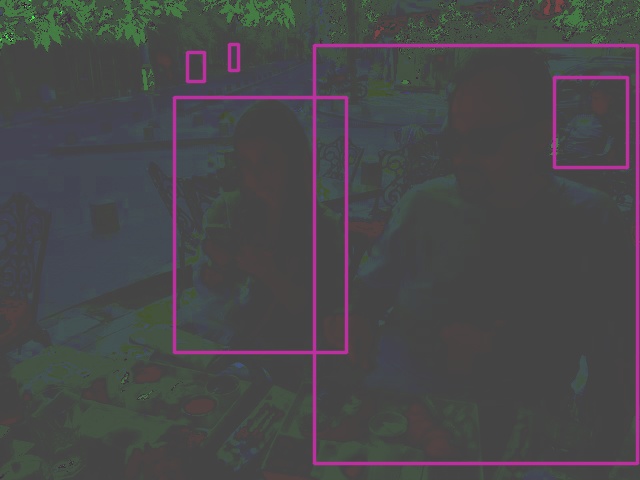}
        \end{minipage}
        \begin{minipage}[t]{0.32\linewidth}
            \centering
            \includegraphics[width=\linewidth,height=0.7\textwidth]{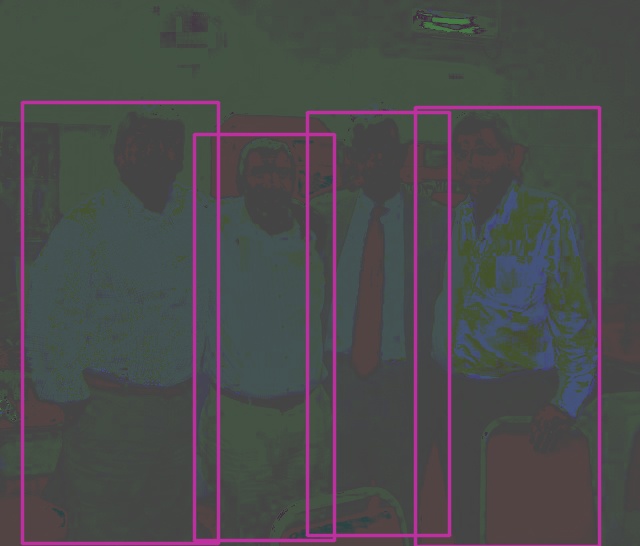}
        \end{minipage}
    }
    \subfigure[\ours enhanced images]{
        \begin{minipage}[t]{0.32\linewidth}
            \centering
            \includegraphics[width=\linewidth,height=0.7\textwidth]{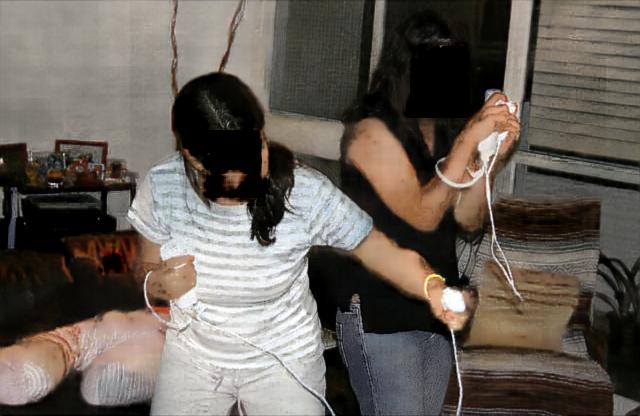}
        \end{minipage}
        \begin{minipage}[t]{0.32\linewidth}
            \centering
            \includegraphics[width=\linewidth,height=0.7\textwidth]{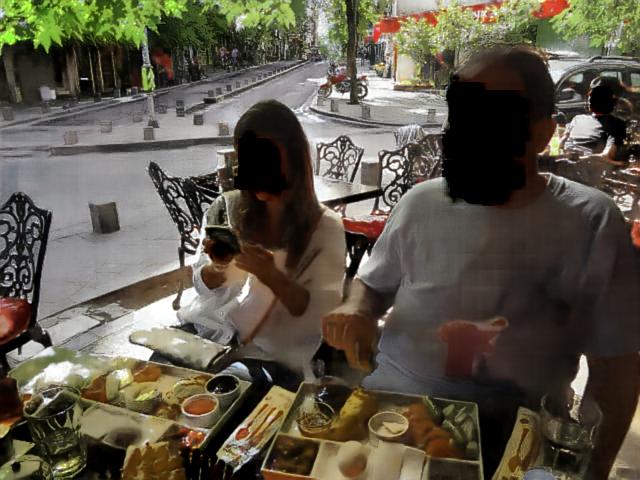}
        \end{minipage}
        \begin{minipage}[t]{0.32\linewidth}
            \centering
            \includegraphics[width=\linewidth,height=0.7\textwidth]{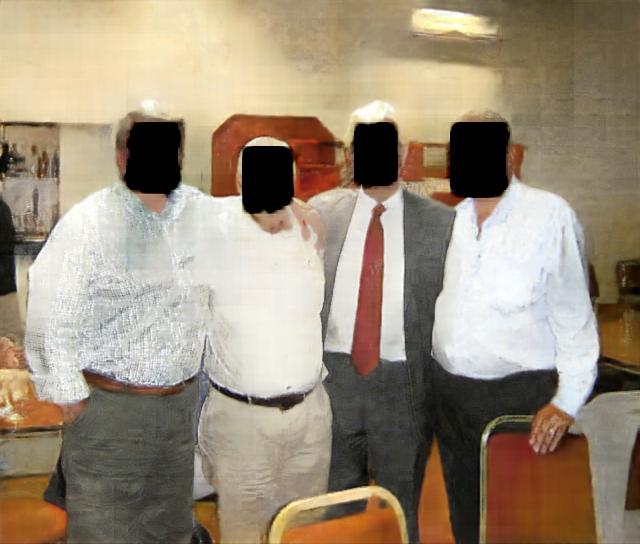}
        \end{minipage}
    }
    \caption{Qualitative results of \ours. Fig.~(a) and Fig.~(b) show the results of person detection on raw images and captured images, respectively. The utility of person detection preserves with \ours. Fig.~(c) shows the enhanced images that are friendly for humans to perceive the human activities.}
    \label{fig:utility}
\end{figure}

We evaluate the person detection results of YOLOv5m, i.e., the targeted vision application, on COCO as the overall performance.
Besides the cases with and without \ours, we investigate two existing hardware-level privacy-preserving approaches, i.e., using a low-resolution camera~\cite{ryoo2017privacy} and using a defocused camera~\cite{Pittaluga2015PrivacyPO}, as the baselines of \ours. To have a fair comparison, we control under a similar degree of privacy protection as \ours. The details of our implementation can be found in Appendix~\ref{app:baseline}.
The quantitative results of person detection are presented in Table~\ref{tab:utility-cap}. The AP, which is the primary challenge metric of COCO, on raw images is 0.578. \ours preserves an AP of 0.475 while the low-resolution and defocused approach obtains a worse AP of 0.284 and 0.395 than \ours, respectively. For other metrics, e.g., AP with different IoU thresholds (AP@0.5, AP@0.75), precision, recall, and F1 score, \ours also performs better than the low-resolution and defocused approach. The qualitative results of the captured images are shown in Fig.~\ref{fig:utility}(b).

\begin{table}[]
    \centering
    \setlength{\tabcolsep}{6pt}
    \caption{Utility of Person Detection}
    \label{tab:utility-cap}
    \begin{tabular}{@{}lcccccc@{}}
        \toprule
        & AP & AP@0.5 & AP@0.75 & Precision & Recall & F1 \\ \midrule
        Raw Images & 0.578 & 0.833 & 0.625 & 0.840 & 0.739 & 0.786 \\
        Low-Resolution & 0.284 & 0.517 & 0.271 & 0.722 & 0.444 & 0.550 \\
        Defocused & 0.395 & 0.655 & 0.399 & 0.780 & 0.565 & 0.655 \\
        \ours & \textbf{0.475} & \textbf{0.742} & \textbf{0.496} & \textbf{0.796} & \textbf{0.650} & \textbf{0.716} \\ \bottomrule
    \end{tabular}
    \vspace{-5pt}
\end{table}

\subsubsection{Image Quality Assessment}
For human perception, we mainly evaluate it from the aspect of image quality. Specifically, we employ the full reference image quality assessment~\cite{hore2010image,wang2004image,wang2003multiscale}, i.e., calculating the similarity of the test image with the reference image. When the test image, i.e., the protected image, is identical to the reference image, i.e., the raw image, the assessed image quality achieves the best. From the results shown in Table~\ref{tab:iqa}, we find that the image enhancement promotes PSNR from 10.8 dB to 21.5 dB, SSIM from 0.437 to 0.749, and MS-SSIM from 0.195 to 0.761. According to prior works~\cite{thomos2005optimized,hameed2021perceptually}, the image quality of the enhanced images is within the acceptable range, i.e., with a PSNR larger than 20 dB and an SSIM larger than 0.7. The qualitative results are illustrated in Fig.~\ref{fig:utility}(c).

\begin{table}[]
    \centering
    \setlength{\tabcolsep}{13pt}
    \caption{Image Quality Assessment}
    \label{tab:iqa}
    \begin{threeparttable}
    \begin{tabular}{@{}lcccc@{}}
    \toprule
    Image Type & RMSE$\ \downarrow$ & PSNR$\ \uparrow$ & SSIM$\ \uparrow$ & MS-SSIM$\ \uparrow$ \\ \midrule
    Captured & 0.299 & 10.8 dB & 0.437 & 0.195 \\
    Enhanced & \textbf{0.093} & \textbf{21.5} dB & \textbf{0.749} & \textbf{0.761} \\ \bottomrule
    \end{tabular}
    \begin{tablenotes}
        \item[1] The symbol $\downarrow$ denotes that a lower value is better for the metric. The symbol $\uparrow$ denotes that a higher value is better for the metric.
    \end{tablenotes}
    \end{threeparttable}
    \vspace{-10pt}
\end{table}

\subsubsection{Generalization on Various Vision Applications}
\label{sec:utilityGeneral}
Since the enhanced images achieve a basic level of image quality, we further study whether the enhanced images can be used for various HAR vision applications besides the targeted one. We investigate the other two vision applications, i.e., human pose estimation and image captioning. The models and metrics for evaluation are presented in Sec.~\ref{sec:setup}. Additionally, we use one classic model per vision application as the baseline, i.e., DeepPose~\cite{toshev2014deeppose} for human pose estimation and Att2in~\cite{rennie2017self} for image caption.
As shown in Table~\ref{tab:utility-enh}, the performances of both human pose estimation and image caption on the enhanced images are better than the baselines, indicating that the degradation of utility is acceptable and can be mitigated by the improvement of the recognition model. The qualitative recognition results on the enhanced images are shown in Fig.~\ref{fig:coco} in Appendix.

\begin{table}[]
    \centering
    \setlength{\tabcolsep}{11pt}
    \caption{Utility of Various Vision Applications}
    \label{tab:utility-enh}
    \begin{tabular}{@{}lcccc@{}}
    \toprule
    Vision Application & Metric & Raw & Baseline & \ours \\ \midrule
    Human Pose Estimation & AP & 0.742 & 0.552 & \textbf{0.554} \\
    Image Caption & CIDEr & 1.334 & 1.114 & \textbf{1.131} \\ \bottomrule
    \end{tabular}
\end{table}

\subsection{Privacy-Utility Trade-offs Analysis}
\label{sec:tradeoffs}

\begin{figure}
    \centering
    \includegraphics[width=.8\linewidth]{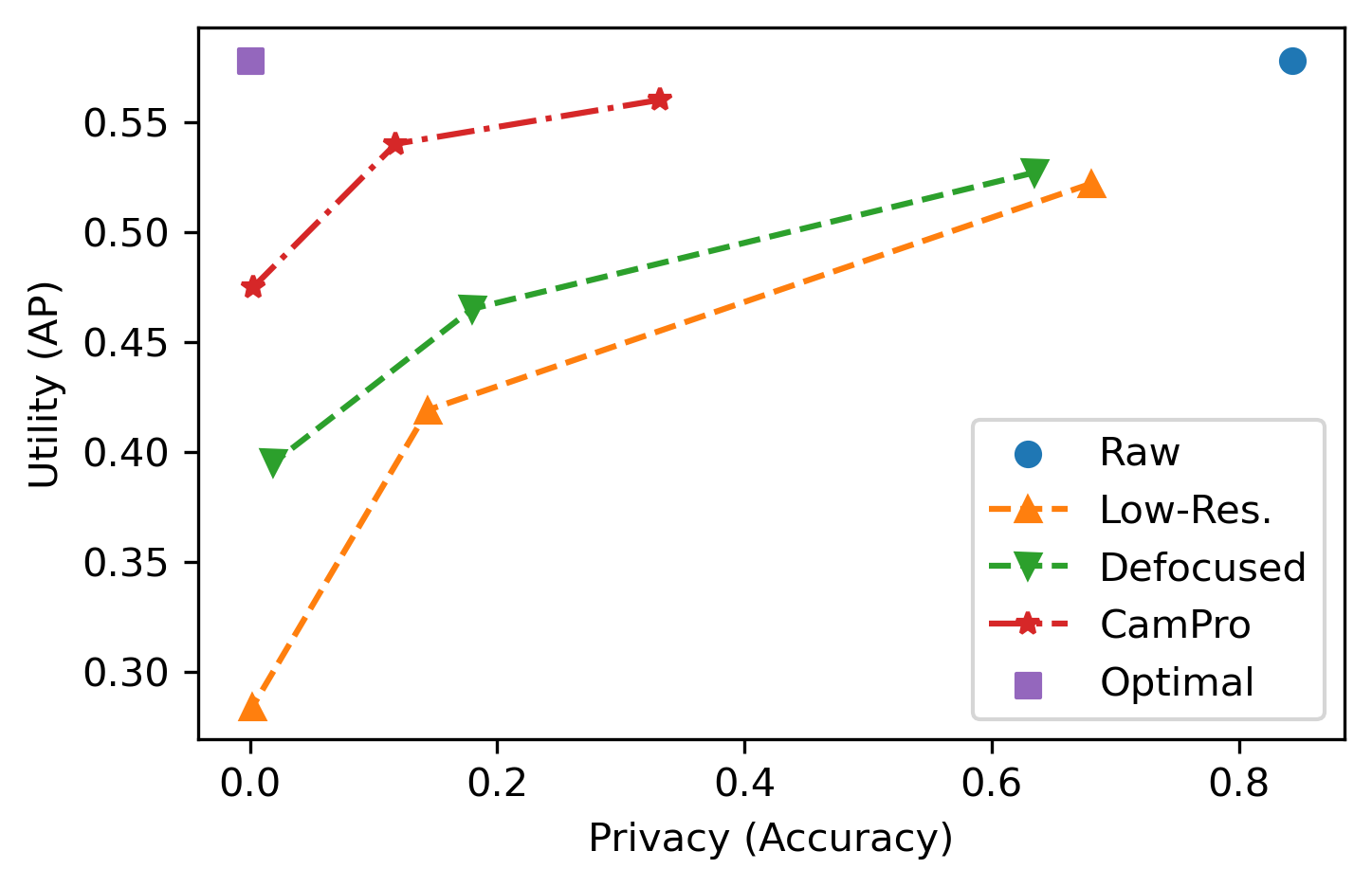}
    \caption{Trade-offs of privacy and utility. The horizontal axis denotes the face identification accuracy, where \textbf{a lower accuracy indicates a better privacy}, and the vertical axis denotes the average precision (AP) of person detection, where \textbf{a higher AP indicates a better utility}. The figure shows that \ours achieves better privacy-utility trade-offs than the baselines, i.e., the low-resolution and defocused camera.}
    \label{fig:baseline}
\end{figure}

Privacy protection and the goal of maintaining utility usually conflict with each other. To investigate the trade-offs of privacy and utility, we change the optimization focus of \ours by adjusting the weight $\omega$ in Eq.~\ref{eq:grad} that balances the privacy loss and the utility loss. As \ours has achieved a strong privacy-preserving effect, i.e., reducing the accuracy to $<1\%$, under the default parameters, we increase the weight $\omega$ to make the optimization more focused on utility. Specifically, we investigate 3 different weights $\omega$, i.e., 0.2 (the default one), 1, and 5. 
For the low-resolution~\cite{ryoo2017privacy} and defocused~\cite{Pittaluga2015PrivacyPO} cameras, we also select 3 groups of parameters that stand for different protection levels per approach, which is presented in Appendix~\ref{app:baseline}. We use the average face identification accuracy of 10 FR models on CelebA to measure the performance of privacy, and utilize the AP of YOLOv5m on COCO to measure the performance of utility. To better understand the trade-offs, we present the performance of privacy and utility in the same coordinate, as shown in Fig.~\ref{fig:baseline}.

From the results, we can see that at the cost of a part of privacy, \ours can maintain more utility, e.g., the AP increases from 0.475 to 0.540, i.e., the drop of AP is limited within 0.03, while the face identification accuracy increases from 0.3\% to 11.8\%. Furthermore, we observe that the two baseline approaches are dominated by \ours according to the Pareto optimality, which indicates \ours can achieve a better privacy-utility trade-off than previous approaches.

\subsection{Real-world Evaluation}

\begin{figure}
    \centering
    \includegraphics[width=.8\linewidth]{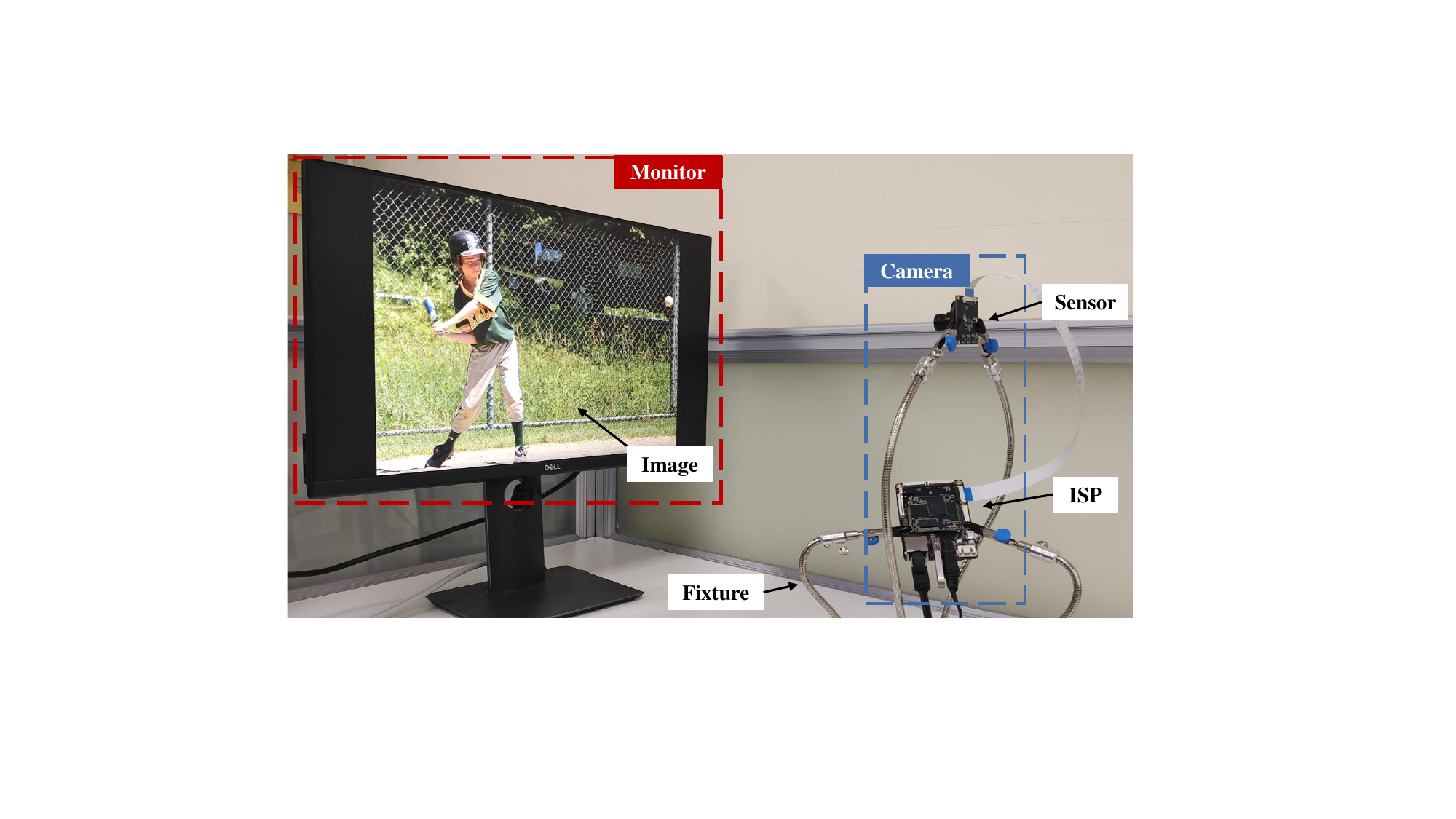}
    \caption[short]{Real-world evaluation setups. We use a camera module, i.e., an ISP RV1126 and an image sensor IMX415, to film the images displayed on a monitor. The \ours parameters are deployed onto the camera module.}
    \label{fig:setups}
\end{figure}

In the real world, we implement a proof-of-concept \ours system using a commodity camera module, with an ISP RV1126~\cite{rv1126} and an image sensor IMX415. We deploy the optimized ISP parameters onto the camera module and implement the image enhancer on a remote server equipped with one NVIDIA RTX 3090 for processing. To ease the burden of recruiting many volunteers for face identification in the real world, we use the camera to film the images displayed on a monitor, as shown in Fig.~\ref{fig:setups}. The captured images are with a resolution of $1920 \times 1080$ and encoded with JPEG format. For each displayed image, we capture twice, where one is captured with the default ISP parameters and the other is captured with the \ours ISP parameters. Due to the overheads to capture images in the real world, we evaluate the effectiveness of \ours on LFW and a subset of COCO that consists of 1,256 images.

\begin{table}[]
    \centering
    \setlength{\tabcolsep}{12pt}
    \caption{Similarity of Image Pairs}
    \label{tab:rw}
    \begin{tabular}{@{}cccccc@{}}
    \toprule
    Image A & Image B & RMSE$\ \downarrow$ & PSNR$\ \uparrow$ & SSIM$\ \uparrow$ \\ \midrule
    simulated & real-world & 0.034 & 30.1 dB & 0.892 \\ 
    raw & enhanced & 0.129 & 17.9 dB & 0.622 \\ 
    \bottomrule
    \end{tabular}
\end{table}

\subsubsection{Image Discrepancy}
\label{sec:realworldimage}
To investigate the discrepancy between simulation and real-world image-taking, we measure the similarities between the simulated captured images and the real-world captured images, as illustrated in Fig.~\ref{fig:rw1} in Appendix. Quantitatively, they achieve a PSNR of 30.1 db, and an SSIM of 0.892, as presented in Table~\ref{tab:rw}. The result implies that our virtual imaging pipeline is fidelity to the real-world image capturing. Moreover, we measure the similarities between the raw images and the enhanced images, i.e., image quality, as shown in Fig.~\ref{fig:rw2} in Appendix. They achieve a PSNR of 17.9 db, and an SSIM of 0.622, which are a bit lower than the ones of simulation due to real-world noises caused by sensor and lossy JPEG compression.

\subsubsection{Performance of Privacy and Utility}
The average accuracy of the 10 face identification systems is 0.13\% on the captured images and 0.28\% on the enhanced images while it is 95.9\% on the raw images. All the accuracies are slightly lower than those in simulation because of real-world noises. The person detector, YOLOv5m, achieves an AP score of 0.648 on the captured images, with respect of the pseudo ground truths that are detection results with high confidence ($>0.5$) on the unprotected images. In general, the real-world performances of privacy and utility are consistent with our simulation results.


\section{Security Analysis}
\label{sec:defense}

In this section, we analyze the security of \ours against adaptive adversaries who are aware of our protection and try to bypass it. From the technical perspective, we investigate two representative kinds of adaptive attacks, i.e., (1) image restoration, and (2) model re-training. The main idea of the former is to improve the data for recognition, i.e., to convert the protected faces into the raw faces with state-of-the-art image restoration methods. The major objective of the latter is to improve the recognition model, i.e., to specialize the recognition model to discriminate the facial features of different subjects shown in protected images. We assume that all the adaptive attacks target at the outputs of camera, i.e., the captured images, rather than the outputs of image enhancer because the facial regions in the enhanced images are completely removed. Moreover, we consider 3 types of adversaries with different levels of prior knowledge, i.e., black-box adversaries, gray-box adversaries, and white-box adversaries. Their capabilities are presented as follows:
\begin{enumerate}
    \item \textit{Black-box adversaries} have no knowledge of \ours. They may apply existing image restoration techniques onto the protected images.
    \item \textit{Gray-box adversaries} have limited knowledge of \ours. They may purchase the same type of camera equipped with \ours and use it to capture images for reference. However, the ISP parameters of \ours are not accessible to them.
    \item \textit{White-box adversaries} have full knowledge of \ours. They may know the optimized ISP parameters of \ours, if possible, via social engineering or reverse-engineering the same type of camera.
\end{enumerate}

\subsection{Black-box Adversary}

\begin{figure}
    \centering
    \subfigbottomskip=-1pt
    \subfigcapskip=-3pt
    \subfigure[Raw Images]{\includegraphics[width=.48\linewidth]{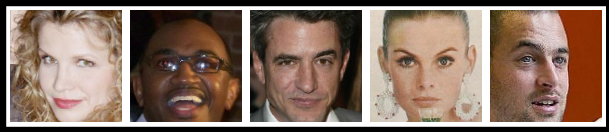}}
    \subfigure[\ours Captured Images]{\includegraphics[width=.48\linewidth]{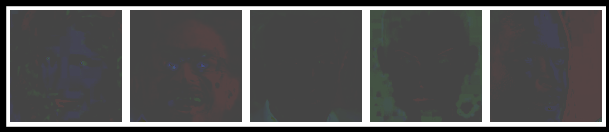}}
    \subfigure[Image Denoising]{\includegraphics[width=.48\linewidth]{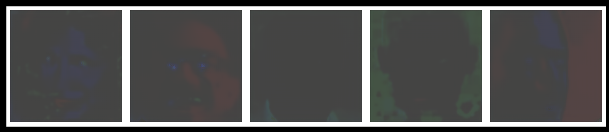}}
    \subfigure[Image Deblurring]{\includegraphics[width=.48\linewidth]{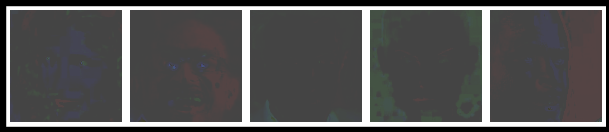}}
    \subfigure[Blind Face Restoration]{\includegraphics[width=.48\linewidth]{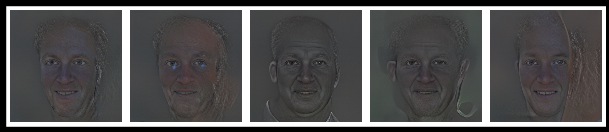}}
    \subfigure[White-box Restoration]{\includegraphics[width=.48\linewidth]{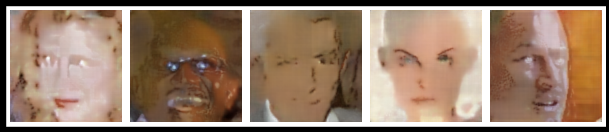}}
    \caption{Qualitative results of adaptive attacks based on image restoration. Fig.~(c)-(f) demonstrate the restored images of various approaches, where the images in Fig.~(c)-(e) are restored without prior knowledge while the results in Fig.~(f) with full knowledge of \ours. The best of them still lose many facial details; hence, \ours can still preserve a certain degree of privacy.}
    \label{fig:defense1}
\end{figure}

For black-box adversaries, they have the same capability as evaluated in Sec.~\ref{sec:privacy}. They can only obtain images via the victim's cameras. It is not feasible for them to re-train the FR models because of the lack of labels. Therefore, we mainly discuss the adaptive attacks based on image restoration for black-box adversaries. On review of existing image restoration approaches, we find that few are well-designed for \ours. We investigate two state-of-the-art DNN-based image restoration approaches, i.e., Uformer~\cite{wang2022uformer} and GFP-GAN~\cite{wang2021towards}. Uformer is a transformer-based encoder-decoder neural network that can be used for both image denoising and deblurring~\cite{wang2022uformer}. GFP-GAN is a blind face restoration approach that utilizes the facial priors in a pretrained face GAN~\cite{wang2021towards}. As shown in Fig.~\ref{fig:defense1}, after denoising and deblurring, the images have few differences from the captured images while after blind face restoration, one clearer face appears on the output image, but apparently the face does not belong to the ground-truth subject. It is not surprising that existing image restoration methods perform poorly, because the color distortion introduced by \ours is very different from the natural distortions, e.g., noise, blur, vintage, etc., for which those methods are designed.

\subsection{Gray-box Adversary}

For gray-box adversaries, they may purchase the same type of camera equipped with \ours. Although both black-box and gray-box adversaries can obtain images via calling the camera, the gray-box ones can utilize the purchased camera to produce self-made images, where the subjects are under control. In this way, they can collect pairs of raw images and protected images theoretically. However, it will be a large overhead to manually collect a huge number of protected images for re-training the FR model in the real world. More practically, the attackers may use the physically-accessible camera equipped with \ours to capture several images of subjects in the gallery set to re-enroll the subject with the private image. We implemented such an operation via simulation, and evaluate its impact on the face identification systems used in Sec.~\ref{sec:privacy}.

From the results, we find that, after re-enrolling, the average face identification accuracy rises from 0.3\% to 0.6\% on CelebA and from 0.6\% to 1.3\% on LFW. The accuracy on the captured images is promoted but even not higher than the naive accuracy on the enhanced images. The results indicate that even if we do not introduce any randomness into the imaging pipeline, the adversary can not attain high accuracy by simply replacing the `template', i.e., the enrolled face, with a transformed version. The primary reason is that the normal FR models are unable to extract distinguishable features from the protected image for the subsequent face identification classifier, which is consistent with our observations on the t-SNE plot of extracted facial features, as presented in Sec.~\ref{sec:feature}.

\subsection{White-box Adversary}

\begin{table}[]
    \centering
\caption{Performance of Privacy Protection in the White-box Setting}
\label{tab:adaptive}
\begin{threeparttable}
\begin{tabular}{@{\extracolsep{7.5pt}}lrrrrr@{}}
\toprule
 & \multicolumn{2}{c}{Finetune} & \multicolumn{2}{c} {Train From Scratch} & \multirow[c]{2}{*}{Restoration}  \\ \cmidrule{2-3} \cmidrule{4-5}
 & Softmax & ArcFace & Softmax & ArcFace &  \\ \midrule
FaceNet\tnote{*} & 12.0\% & 0.0\% & 2.3\% & 0.0\% & 2.1\%  \\
Arc18\tnote{*} & 10.1\% & 15.4\% & 6.2\% & 4.7\% & 2.1\%  \\
Arc50\tnote{*} & 19.5\% & 0.0\% & 4.1\% & 10.7\% & 4.7\%  \\
Arc152\tnote{*} & 3.7\% & 0.0\% & 12.6\% & 9.3\% & 3.9\%  \\
Mag18\tnote{*} & 14.5\% & 18.7\% & 7.1\% & 5.7\% & 2.1\%  \\
Mag50\tnote{*} & 15.6\% & 0.0\% & 8.0\% & 0.0\% & 6.3\%  \\
Mag100\tnote{*} & 6.9\% & 0.0\% & 5.3\% & 0.0\% & 7.5\%  \\
Ada18\tnote{*} & 5.4\% & 11.8\% & 3.0\% & 5.3\% & 5.4\%  \\
Ada50\tnote{*} & 18.9\% & 10.1\% & 5.8\% & 13.2\% & 8.3\%  \\
Ada100\tnote{*} & 5.0\% & 10.9\% & 2.1\% & 8.5\% & 10.2\%  \\ \midrule
Average & 11.2\% & 6.7\% & 5.7\% & 5.7\% & 5.3\%  \\ \bottomrule
\end{tabular}
\begin{tablenotes}
    \item[*] The abbreviations are consistent with those shown in Table~\ref{tab:privacy}.
\end{tablenotes}
\end{threeparttable}
\vspace{-5pt}
\end{table}

For white-box adversaries, they may know the built-in parameters of \ours by social engineering or reverse-engineering the same type of camera equipped with \ours. In this way, they can constitute a simulation pipeline to generate numerous images of \ours at a low cost. 
In the following, we consider two types of adaptive attacks, i.e., (1) re-training the facial feature extractor, and (2) developing a specialized image restoration algorithm.

\subsubsection{Model Re-training}
We construct a \ours facial image dataset CelebA-P by converting the normal facial images into the simulated captured images by \ours using the proposed virtual imaging pipeline. Then, we re-train the 10 FR models on CelebA-P by 2 common modes, i.e., training-from-scratch and finetuning. In the training-from-scratch, the weights of the FR models are initialized randomly, while in the finetuning, they are initialized with the pre-trained weights. Moreover, we investigate 2 types of training loss, i.e., Softmax loss and Additive Angular Margin (ArcFace) loss. Therefore, we re-train 40 ($10\times2\times2$) models in total. The model training proceeds 20 epochs. We use an SGD optimizer with a learning rate of 1e-1 and a weight decay of 5e-4. The learning rate of SGD decays to 1e-2 at the end of epoch 15. The above training settings are validated by training with normal face datasets.

\begin{figure}
    \centering
    \includegraphics[width=.95\linewidth]{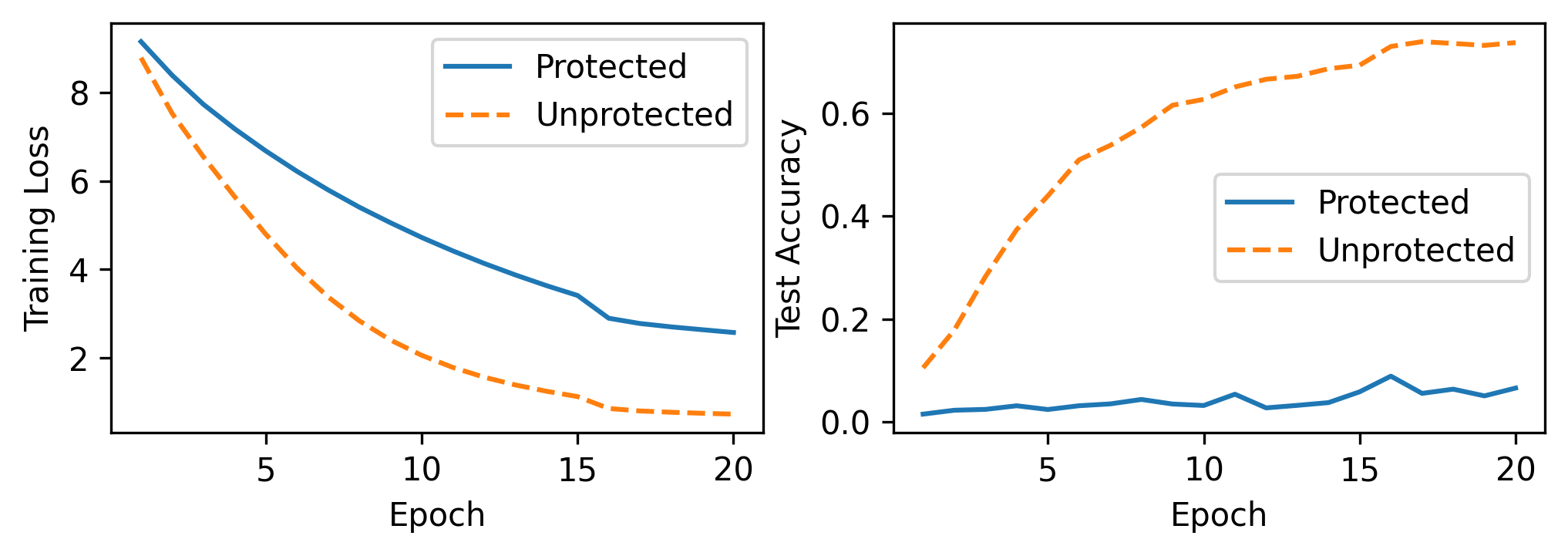}
    \vspace{-4pt}
    \caption{Overfitting phenomenon during model re-training. The training losses of both protected and unprotected images steadily decrease, but the test accuracy of protected images oscillates and stays at a low level.}
    \vspace{-10pt}
    \label{fig:overfit}
\end{figure}

We present the face identification accuracies on CelebA-P in Table~\ref{tab:adaptive} Column 1-4. The highest accuracy among 40 models is 19.5\%, obtained by the ArcFace-IResNet50 finetuned with the Softmax loss. In general, the results of finetuning perform better than the ones of training-from-scratch, and surprisingly, the results of Softmax perform better than the ones of ArcFace which often performs better than Softmax on normal face datasets. Furthermore, the results have a large variance, especially for those using ArcFace loss. That is because the overfitting phenomenon often occurs when training with protected images, as shown in Fig.~\ref{fig:overfit}. The main reason for overfitting is that the training data, i.e., protected images, are highly noisy, indicating that the majority of facial features have been removed by \ours.

\subsubsection{Image Restoration}
Since \ours gives a many-to-one color mapping, it is an ill-posed problem to restore the real-world scene with the captured image. However, neural networks excel at identifying patterns in the degraded image and estimating the original image by being trained on datasets of both original images and protected images. The training settings of image restoration are the same as training the enhancer of \ours, as presented in Sec.~\ref{sec:setup}, except for removing the penalty of face restoration.

As demonstrated in Fig.~\ref{fig:defense1}, the results of the specialized image restoration algorithm for \ours are much better than those of existing image restoration methods. The restored faces look similar to the original ones but lack facial details. Moreover, we investigate the face identification accuracy with the restored facial images. The quantitative results are shown in Table~\ref{tab:adaptive} Column 5. The average accuracy is 5.3\%, and the best accuracy among 10 models is 10.2\%, which are more stable but generally worse than those adaptive attacks based on model re-training.

\noindent \textbf{Takeaways.} From the quantitative results shown in Table~\ref{tab:adaptive}, we find that in the face of the strongest white-box adaptive attacks, \ours can lower the average face identification accuracy below 12\% on CelebA. The result indicates that \ours can still preserve a certain degree of privacy, even if the adversary has complete knowledge of \ours.

\section{Limitation \& Discussion}

\noindent \textbf{Limitations.} This study has two limitations. (1) \ours system is not suitable for scenarios that require the fidelity of the user's face, such as taking a selfie, because it is conflicted with the goal of achieving AFR against both automatic programs and humans. (2) The camera modeling of \ours is based on a mainstream architecture of ISP but does not cover all ISP architectures, e.g., those where Gamma is not implemented as LUT. Although the camera modeling need to be replaced according to the ISP architecture, the design of the adversarial learning framework and the image enhancer is still applicable.

\noindent \textbf{Extensive Effects on Face Detection.} In the above evaluation, we suppose that the users' faces protected by \ours are correctly detected as well as the regular faces. However, \ours also affects the accuracy of face detection, which is an essential pre-processing for face identification. If no faces are detected, the procedure of face identification will be aborted. Therefore, \ours may succeed beyond expectation due to its effects on face detection. For instance, we tested 300 \ours images on the face searching (identification) API of Face++~\cite{facepp}, and found that none of the protected faces is recognized by the Face++ API. We also observed frequent failures of face detection on another commercial FR API, Amazon Rekognition~\cite{amazon}.

\noindent \textbf{Deployment on Android.} We attempt to deploy on smartphones, thus preventing unauthorized FR when using those untrusted apps that access the camera. As we do not have the same privilege as the Original Equipment Manufacturers (OEMs) of the smartphones, we failed to modify the interfaces or drivers, where \ours should be deployed, in the firmware. However, we can validate the capabilities of \ours with the interfaces of the Android camera subsystem~\cite{androidHal}. Specifically, we implemented \ours by setting two parameters of the Camera2 API, i.e., \textit{ColorSpaceTransform}~\cite{androidccm} and \textit{TonemapCurve}~\cite{androidgamma}. We investigated 8 Android smartphones with different models, and all of them succeeded to perform the effects of \ours. The specifications of the tested smartphone are presented in Table~\ref{tab:phone} in Appendix.

\section{Related Work}

In this section, we introduce the prior work on anti-facial recognition (AFR) related to \ours. We classify the existing related literatures into two groups by their positions relative to the output images of the camera module, i.e., post-processing-based AFR and pre-processing-based AFR. 

\subsection{Post-Processing-Based Anti-Facial Recognition}

Much work has studied the feasibility of achieving AFR via image post-processing. 
The most classic and general approach is to first detect the faces and then employ simple degradation functions, e.g., pixelating, blurring, and masking, to remove the facial information~\cite{Yang2022ASO,Ilia2015FaceOffPP,Koshimizu2006FactorsOT}.
A branch of work proposes to substitute the simple degradation function to improve the visual experience of human viewers. 
\citeauthor{Bitouk2008FaceSA}~\cite{Bitouk2008FaceSA} propose a complete system for automatic face replacement in images and generate plausible results of pose, lighting, and skin color. 
\citeauthor{Rhee2013CartoonlikeAG}~\cite{Rhee2013CartoonlikeAG} build a system that generates a cartoon character looking similar to the user by compositing the most similar facial components according to facial features. 
\citeauthor{Kuang2021EffectiveDG}~\cite{Kuang2021EffectiveDG} develop a GAN-based approach that replaces the original face with a different yet realistic face via image synthesis. 
Recently, adversarial example attacks~\cite{Goodfellow2015ExplainingAH,Madry2018TowardsDL} are applied to prevent unauthorized FR, e.g., Fawkes~\cite{shan2020fawkes}, LowKey~\cite{cherepanova2021lowkey}, and TIP-IM~\cite{Yang2021TowardsFE}, which achieve minimal modifications to the original image.
Another branch of work studies a similar problem as the one in this paper, i.e., how to preserve useful information but eliminates sensitive information for a given vision application.
\citeauthor{Bertrn2019AdversariallyLR}~\cite{Bertrn2019AdversariallyLR} propose an information theoretic approach based on adversarial games of DNNs to maintain performance on existing algorithms while minimizing sensitive information leakage. 
\citeauthor{Wu2018TowardsPV}~\cite{Wu2018TowardsPV,Wu2020PrivacyPreservingDA} propose a DNN with adversarial learning to apply a degradation transform for the video inputs to make action recognition feasible while suppressing the privacy breach risk.
Although their motivation is similar to ours, all of them use DNN, which is impossible to realize in most camera modules, to desensitize the images as post-processing. In contrast, we achieve the goal with the common camera module by using the built-in functions of its ISP.

\subsection{Pre-Processing-Based Anti-Facial Recognition}

Another group of related work studies privacy preservation with a pre-processing. We place \ours into the pre-processing-based methods that achieve desensitization before the camera module outputs the captured image.
\citeauthor{Pittaluga2015PrivacyPO}~\cite{Pittaluga2015PrivacyPO} design two different optical systems to protect privacy by altering the injected lights of the camera. They perform optical averaging to achieve a K-anonymity image capture, and they use a defocused lens to obfuscate the images captured by infrared cameras, which still enables accurate depth sensing and motion tracking. 
\citeauthor{Hinojosa2021LearningPO}~\cite{Hinojosa2021LearningPO} further dedicate the defocused lens by optimizing the point spread function to prevent FR while maintaining the utility of human pose estimation. A few similar approaches as~\cite{Hinojosa2021LearningPO} can achieve privacy-preserving image caption~\cite{arguello2022optics} and action recognition~\cite{hinojosa2022privhar}. 
In addition to optics, \citeauthor{ryoo2017privacy}~\cite{ryoo2017privacy} propose to perform person detection with an extremely low-resolution camera, i.e., 10x or 20x lower resolution w.r.t. its original resolution, thus making FR difficult.
\citeauthor{Pittaluga2016SensorlevelPF}~\cite{Pittaluga2016SensorlevelPF} design a specialized circuit that utilize the human body temperature to locate and mask the faces inside the thermal camera.
\citeauthor{Wang2022CamShieldSS}~\cite{Wang2022CamShieldSS} propose a physically-isolated shielding system consisting of a camera and a screen, which is placed in front of the original camera to encrypt the face before transmitting it to the Internet. 
Compared to the existing pre-processing-based approaches, \ours dives into the camera module to use its built-in ISP functions, which requires no additional hardware and realizes a tight binding of image acquisition and privacy protection.

\section{Conclusion}

In this paper, we investigate the feasibility of achieving anti-facial recognition (AFR) within a camera module meanwhile maintaining the utility in the context of human activity recognition. We propose \ours, which contains a camera module and an image enhancer, which requires no additional hardware and is practical to deploy on commodity camera modules. The AFR effects of \ours can generalize well on various black-box face identification systems, i.e., reducing the average accuracy to 0.3\%, and are resistant to white-box adaptive attacks. \ours can achieve better privacy-utility trade-offs than previous hardware-level approaches. This work serves as the first attempt at \emph{privacy preservation by birth} leveraging built-in ISP functions in the camera module. Future directions include (1) investigating other ISP functions besides CCM and Gamma, (2) investigating other kinds of image privacy protection besides AFR, and (3) further improving the visual appearance of the output images, possibly with the help of advanced image generation techniques.

\section*{Acknowledgement}

This paper is supported by China NSFC Grant 62222114, 61925109, 62071428, 62271280 and the Fundamental Research Funds for the Central Universities 226-2022-00223.

\bibliographystyle{plainnat}
\bibliography{reference.bib}

\begin{thebibliography}{85}
\providecommand{\natexlab}[1]{#1}
\providecommand{\url}[1]{\texttt{#1}}
\expandafter\ifx\csname urlstyle\endcsname\relax
  \providecommand{\doi}[1]{doi: #1}\else
  \providecommand{\doi}{doi: \begingroup \urlstyle{rm}\Url}\fi

\bibitem[alfa ai(2023)]{aifitness}
alfa ai.
\newblock Ai movements scanning.
\newblock \url{https://www.alfa-ai.com/}, 2023.

\bibitem[Anderson et~al.(1996)Anderson, Motta, Chandrasekar, and
  Stokes]{anderson1996proposal}
Matthew Anderson, Ricardo Motta, Srinivasan Chandrasekar, and Michael Stokes.
\newblock Proposal for a standard default color space for the internet—srgb.
\newblock In \emph{Color and imaging conference}, 1996.

\bibitem[Arguello et~al.(2022)Arguello, Lopez, Hinojosa, and
  Arguello]{arguello2022optics}
Paula Arguello, Jhon Lopez, Carlos Hinojosa, and Henry Arguello.
\newblock Optics lens design for privacy-preserving scene captioning.
\newblock In \emph{2022 IEEE International Conference on Image Processing
  (ICIP)}, 2022.

\bibitem[Bertr{\'a}n et~al.(2019)Bertr{\'a}n, Mart{\'i}nez, Papadaki, Qiu,
  Rodrigues, Reeves, and Sapiro]{Bertrn2019AdversariallyLR}
Mart{\'i}n Bertr{\'a}n, Natalia Mart{\'i}nez, Afroditi Papadaki, Qiang Qiu,
  Miguel R.~D. Rodrigues, Galen Reeves, and Guillermo Sapiro.
\newblock Adversarially learned representations for information obfuscation and
  inference.
\newblock In \emph{International Conference on Machine Learning}, 2019.

\bibitem[Bitouk et~al.(2008)Bitouk, Kumar, Dhillon, Belhumeur, and
  Nayar]{Bitouk2008FaceSA}
Dmitri Bitouk, Neeraj Kumar, Samreen Dhillon, Peter~N. Belhumeur, and Shree~K.
  Nayar.
\newblock Face swapping: automatically replacing faces in photographs.
\newblock \emph{ACM SIGGRAPH 2008 papers}, 2008.

\bibitem[Calipsa(2021)]{Calipsa}
Calipsa.
\newblock The story of cctv in europe, from resistance to adoption.
\newblock \url{https://tinyurl.com/bddwnj6c}, 2021.

\bibitem[Carlini and Wagner(2017)]{carlini2017towards}
Nicholas Carlini and David Wagner.
\newblock Towards evaluating the robustness of neural networks.
\newblock In \emph{2017 ieee symposium on security and privacy (sp)}, 2017.

\bibitem[Chen et~al.(2021{\natexlab{a}})Chen, Zhang, Yao, Guo, Yu, and
  Liu]{chen2021deep}
Kaixuan Chen, Dalin Zhang, Lina Yao, Bin Guo, Zhiwen Yu, and Yunhao Liu.
\newblock Deep learning for sensor-based human activity recognition: Overview,
  challenges, and opportunities.
\newblock \emph{ACM Computing Surveys (CSUR)}, 2021{\natexlab{a}}.

\bibitem[Chen et~al.(2021{\natexlab{b}})Chen, Du, Luo, and
  Xiao]{Chen2021FallDS}
Yang Chen, Rongxi Du, Kaitao Luo, and Yu~Lian Xiao.
\newblock Fall detection system based on real-time pose estimation and svm.
\newblock \emph{2021 IEEE 2nd International Conference on Big Data, Artificial
  Intelligence and Internet of Things Engineering (ICBAIE)},
  2021{\natexlab{b}}.

\bibitem[Cherepanova et~al.(2021)Cherepanova, Goldblum, Foley, Duan, Dickerson,
  Taylor, and Goldstein]{cherepanova2021lowkey}
Valeriia Cherepanova, Micah Goldblum, Harrison Foley, Shiyuan Duan, John~P
  Dickerson, Gavin Taylor, and Tom Goldstein.
\newblock Lowkey: Leveraging adversarial attacks to protect social media users
  from facial recognition.
\newblock In \emph{Proceedings of the International Conference on Learning
  Representations (ICLR)}, 2021.

\bibitem[Contributors(2020)]{mmpose2020}
MMPose Contributors.
\newblock Openmmlab pose estimation toolbox and benchmark.
\newblock \url{https://github.com/open-mmlab/mmpose}, 2020.

\bibitem[Dang et~al.(2020)Dang, Min, Wang, Piran, Lee, and
  Moon]{dang2020sensor}
L~Minh Dang, Kyungbok Min, Hanxiang Wang, Md~Jalil Piran, Cheol~Hee Lee, and
  Hyeonjoon Moon.
\newblock Sensor-based and vision-based human activity recognition: A
  comprehensive survey.
\newblock \emph{Pattern Recognition}, 2020.

\bibitem[Deng et~al.(2019)Deng, Guo, Xue, and Zafeiriou]{deng2019arcface}
Jiankang Deng, Jia Guo, Niannan Xue, and Stefanos Zafeiriou.
\newblock Arcface: Additive angular margin loss for deep face recognition.
\newblock In \emph{Proceedings of the IEEE/CVF conference on computer vision
  and pattern recognition}, 2019.

\bibitem[Deng et~al.(2020)Deng, Guo, Ververas, Kotsia, Zafeiriou, and
  FaceSoft]{Deng2020RetinaFaceSM}
Jiankang Deng, J.~Guo, Evangelos Ververas, Irene Kotsia, Stefanos Zafeiriou,
  and InsightFace FaceSoft.
\newblock Retinaface: Single-shot multi-level face localisation in the wild.
\newblock \emph{2020 IEEE/CVF Conference on Computer Vision and Pattern
  Recognition (CVPR)}, 2020.

\bibitem[Du et~al.(2022)Du, Shi, Zeng, Zhang, and Mei]{du2022elements}
Hang Du, Hailin Shi, Dan Zeng, Xiao-Ping Zhang, and Tao Mei.
\newblock The elements of end-to-end deep face recognition: A survey of recent
  advances.
\newblock \emph{ACM Computing Surveys (CSUR)}, 2022.

\bibitem[Face++(2022)]{facepp}
Face++.
\newblock Face searching.
\newblock \url{https://www.faceplusplus.com/face-searching/}, 2022.

\bibitem[Finlayson et~al.(2015)Finlayson, Mackiewicz, and
  Hurlbert]{finlayson2015color}
Graham~D Finlayson, Michal Mackiewicz, and Anya Hurlbert.
\newblock Color correction using root-polynomial regression.
\newblock \emph{IEEE Transactions on Image Processing}, 2015.

\bibitem[Fraden and Fraden(2010)]{fraden2010handbook}
Jacob Fraden and Jacob Fraden.
\newblock \emph{Handbook of modern sensors: physics, designs, and
  applications}, volume~3.
\newblock Springer, 2010.

\bibitem[Fulton(2022)]{gamma}
Wayne Fulton.
\newblock What and why is gamma correction in photo images?
\newblock \url{https://www.scantips.com/lights/gamma2.html}, 2022.

\bibitem[Goodfellow et~al.(2014)Goodfellow, Pouget-Abadie, Mirza, Xu,
  Warde-Farley, Ozair, Courville, and Bengio]{Goodfellow2014GenerativeAN}
Ian~J. Goodfellow, Jean Pouget-Abadie, Mehdi Mirza, Bing Xu, David
  Warde-Farley, Sherjil Ozair, Aaron~C. Courville, and Yoshua Bengio.
\newblock Generative adversarial nets.
\newblock In \emph{NIPS}, 2014.

\bibitem[Goodfellow et~al.(2015)Goodfellow, Shlens, and
  Szegedy]{Goodfellow2015ExplainingAH}
Ian~J. Goodfellow, Jonathon Shlens, and Christian Szegedy.
\newblock Explaining and harnessing adversarial examples.
\newblock \emph{CoRR}, abs/1412.6572, 2015.

\bibitem[Google(2022)]{androidHal}
Google.
\newblock The hal and camera subsystem.
\newblock \url{https://tinyurl.com/te8kmj3u}, 2022.

\bibitem[Google(2023{\natexlab{a}})]{androidccm}
Google.
\newblock Color space transform.
\newblock \url{https://tinyurl.com/3xjuftxj}, 2023{\natexlab{a}}.

\bibitem[Google(2023{\natexlab{b}})]{androidgamma}
Google.
\newblock Tonemap curve.
\newblock \url{https://tinyurl.com/46uujf8z}, 2023{\natexlab{b}}.

\bibitem[G{\"u}nther et~al.(2017)G{\"u}nther, Cruz, Rudd, and
  Boult]{Gnther2017TowardOF}
Manuel G{\"u}nther, Steve Cruz, Ethan~M. Rudd, and Terrance~E. Boult.
\newblock Toward open-set face recognition.
\newblock \emph{2017 IEEE Conference on Computer Vision and Pattern Recognition
  Workshops (CVPRW)}, 2017.

\bibitem[Hameed and Gyorgy(2021)]{hameed2021perceptually}
Muhammad~Zaid Hameed and Andras Gyorgy.
\newblock Perceptually constrained adversarial attacks.
\newblock \emph{arXiv preprint arXiv:2102.07140}, 2021.

\bibitem[Hatmaker(2020)]{hatmaker2020portland}
Taylor Hatmaker.
\newblock Portland passes expansive city ban on facial recognition tech.
\newblock \emph{TechCrunch.[Google Scholar]}, 2020.

\bibitem[Hinojosa et~al.(2021)Hinojosa, Niebles, and
  Arguello]{Hinojosa2021LearningPO}
Carlos Hinojosa, Juan~Carlos Niebles, and Henry Arguello.
\newblock Learning privacy-preserving optics for human pose estimation.
\newblock \emph{2021 IEEE/CVF International Conference on Computer Vision
  (ICCV)}, 2021.

\bibitem[Hinojosa et~al.(2022)Hinojosa, Marquez, Arguello, Adeli, Fei-Fei, and
  Niebles]{hinojosa2022privhar}
Carlos Hinojosa, Miguel Marquez, Henry Arguello, Ehsan Adeli, Li~Fei-Fei, and
  Juan~Carlos Niebles.
\newblock Privhar: Recognizing human actions from privacy-preserving lens.
\newblock In \emph{Computer Vision--ECCV 2022: 17th European Conference, Tel
  Aviv, Israel, October 23--27, 2022, Proceedings, Part IV}, 2022.

\bibitem[Hore and Ziou(2010)]{hore2010image}
Alain Hore and Djemel Ziou.
\newblock Image quality metrics: Psnr vs. ssim.
\newblock In \emph{2010 20th international conference on pattern recognition},
  2010.

\bibitem[Huang et~al.(2008)Huang, Mattar, Berg, and
  Learned-Miller]{huang2008labeled}
Gary~B Huang, Marwan Mattar, Tamara Berg, and Eric Learned-Miller.
\newblock Labeled faces in the wild: A database forstudying face recognition in
  unconstrained environments.
\newblock In \emph{Workshop on faces in'Real-Life'Images: detection, alignment,
  and recognition}, 2008.

\bibitem[Ilia et~al.(2015)Ilia, Polakis, Athanasopoulos, Maggi, and
  Ioannidis]{Ilia2015FaceOffPP}
Panagiotis Ilia, Iasonas Polakis, Elias Athanasopoulos, Federico Maggi, and
  Sotiris Ioannidis.
\newblock Face/off: Preventing privacy leakage from photos in social networks.
\newblock \emph{Proceedings of the 22nd ACM SIGSAC Conference on Computer and
  Communications Security}, 2015.

\bibitem[Joe(2021)]{joe2021tiktok}
Walsh Joe.
\newblock Tiktok settles privacy lawsuit for \$92 million.
\newblock \emph{Forbes}, 2021.

\bibitem[Kemelmacher-Shlizerman et~al.(2015)Kemelmacher-Shlizerman, Seitz,
  Miller, and Brossard]{KemelmacherShlizerman2015TheMB}
Ira Kemelmacher-Shlizerman, Steven~M. Seitz, Daniel Miller, and Evan Brossard.
\newblock The megaface benchmark: 1 million faces for recognition at scale.
\newblock \emph{2016 IEEE Conference on Computer Vision and Pattern Recognition
  (CVPR)}, 2015.

\bibitem[Kim et~al.(2022)Kim, Jain, and Liu]{Kim2022AdaFaceQA}
Minchul Kim, Anil~K. Jain, and Xiaoming Liu.
\newblock Adaface: Quality adaptive margin for face recognition.
\newblock \emph{2022 IEEE/CVF Conference on Computer Vision and Pattern
  Recognition (CVPR)}, 2022.

\bibitem[Kingma and Ba(2014)]{Kingma2014AdamAM}
Diederik~P. Kingma and Jimmy Ba.
\newblock Adam: A method for stochastic optimization.
\newblock \emph{CoRR}, abs/1412.6980, 2014.

\bibitem[Koshimizu et~al.(2006)Koshimizu, Toriyama, and
  Babaguchi]{Koshimizu2006FactorsOT}
Takashi Koshimizu, Tomoji Toriyama, and Noboru Babaguchi.
\newblock Factors on the sense of privacy in video surveillance.
\newblock In \emph{CARPE '06}, 2006.

\bibitem[Kuang et~al.(2021)Kuang, Liu, Yu, Tian, Wang, Fan, and
  Babaguchi]{Kuang2021EffectiveDG}
Zhenzhong Kuang, Huigui Liu, Jun Yu, Aikui Tian, Lei Wang, Jianping Fan, and
  Noboru Babaguchi.
\newblock Effective de-identification generative adversarial network for face
  anonymization.
\newblock \emph{Proceedings of the 29th ACM International Conference on
  Multimedia}, 2021.

\bibitem[Li et~al.(2022{\natexlab{a}})Li, Li, Le, Wang, Savarese, and
  Hoi]{li2022lavis}
Dongxu Li, Junnan Li, Hung Le, Guangsen Wang, Silvio Savarese, and Steven C.~H.
  Hoi.
\newblock Lavis: A library for language-vision intelligence,
  2022{\natexlab{a}}.

\bibitem[Li et~al.(2022{\natexlab{b}})Li, Li, Xiong, and Hoi]{Li2022BLIPBL}
Junnan Li, Dongxu Li, Caiming Xiong, and Steven C.~H. Hoi.
\newblock Blip: Bootstrapping language-image pre-training for unified
  vision-language understanding and generation.
\newblock In \emph{International Conference on Machine Learning},
  2022{\natexlab{b}}.

\bibitem[Lin et~al.(2014)Lin, Maire, Belongie, Hays, Perona, Ramanan,
  Doll{\'a}r, and Zitnick]{lin2014microsoft}
Tsung-Yi Lin, Michael Maire, Serge Belongie, James Hays, Pietro Perona, Deva
  Ramanan, Piotr Doll{\'a}r, and C~Lawrence Zitnick.
\newblock Microsoft coco: Common objects in context.
\newblock In \emph{European conference on computer vision}, 2014.

\bibitem[Liu et~al.(2015)Liu, Luo, Wang, and Tang]{liu2015faceattributes}
Ziwei Liu, Ping Luo, Xiaogang Wang, and Xiaoou Tang.
\newblock Deep learning face attributes in the wild.
\newblock In \emph{Proceedings of International Conference on Computer Vision
  (ICCV)}, 2015.

\bibitem[Loshchilov and Hutter(2017)]{Loshchilov2017DecoupledWD}
Ilya Loshchilov and Frank Hutter.
\newblock Decoupled weight decay regularization.
\newblock In \emph{International Conference on Learning Representations}, 2017.

\bibitem[Madry et~al.(2018)Madry, Makelov, Schmidt, Tsipras, and
  Vladu]{Madry2018TowardsDL}
Aleksander Madry, Aleksandar Makelov, Ludwig Schmidt, Dimitris Tsipras, and
  Adrian Vladu.
\newblock Towards deep learning models resistant to adversarial attacks.
\newblock \emph{ArXiv}, abs/1706.06083, 2018.

\bibitem[Meng et~al.(2021)Meng, Zhao, Huang, and Zhou]{Meng2021MagFaceAU}
Qiang Meng, Shichao Zhao, Zhida Huang, and Feng Zhou.
\newblock Magface: A universal representation for face recognition and quality
  assessment.
\newblock \emph{2021 IEEE/CVF Conference on Computer Vision and Pattern
  Recognition (CVPR)}, 2021.

\bibitem[Nikouei et~al.(2018)Nikouei, Chen, Song, Xu, Choi, and
  Faughnan]{nikouei2018smart}
Seyed~Yahya Nikouei, Yu~Chen, Sejun Song, Ronghua Xu, Baek-Young Choi, and
  Timothy Faughnan.
\newblock Smart surveillance as an edge network service: From harr-cascade, svm
  to a lightweight cnn.
\newblock In \emph{2018 ieee 4th international conference on collaboration and
  internet computing (cic)}, 2018.

\bibitem[of~Standards and (NIST)(2023{\natexlab{a}})]{frvt}
National~Institute of~Standards and Technology (NIST).
\newblock Frvt 1:n identification.
\newblock \url{https://pages.nist.gov/frvt/html/frvt1N.html},
  2023{\natexlab{a}}.

\bibitem[of~Standards and (NIST)(2023{\natexlab{b}})]{pii}
National~Institute of~Standards and Technology (NIST).
\newblock personally identifiable information.
\newblock \url{https://tinyurl.com/3jcmdbku}, 2023{\natexlab{b}}.

\bibitem[Pardau(2018)]{pardau2018california}
Stuart~L Pardau.
\newblock The california consumer privacy act: Towards a european-style privacy
  regime in the united states.
\newblock \emph{J. Tech. L. \& Pol'y}, 2018.

\bibitem[Patrick(2021)]{patrick2021historic}
McKnight Patrick.
\newblock Historic biometric privacy suit settles for \$650 million.
\newblock \emph{Business Law Today}, 2021.

\bibitem[Pittaluga and Koppal(2015)]{Pittaluga2015PrivacyPO}
F.~Pittaluga and Sanjeev~J. Koppal.
\newblock Privacy preserving optics for miniature vision sensors.
\newblock \emph{2015 IEEE Conference on Computer Vision and Pattern Recognition
  (CVPR)}, 2015.

\bibitem[Pittaluga et~al.(2016)Pittaluga, Zivkovic, and
  Koppal]{Pittaluga2016SensorlevelPF}
F.~Pittaluga, Aleksandar Zivkovic, and Sanjeev~J. Koppal.
\newblock Sensor-level privacy for thermal cameras.
\newblock \emph{2016 IEEE International Conference on Computational Photography
  (ICCP)}, 2016.

\bibitem[Punn et~al.(2020)Punn, Sonbhadra, Agarwal, and
  Rai]{punn2020monitoring}
Narinder~Singh Punn, Sanjay~Kumar Sonbhadra, Sonali Agarwal, and Gaurav Rai.
\newblock Monitoring covid-19 social distancing with person detection and
  tracking via fine-tuned yolo v3 and deepsort techniques.
\newblock \emph{arXiv preprint arXiv:2005.01385}, 2020.

\bibitem[Radiya-Dixit et~al.(2022)Radiya-Dixit, Hong, Carlini, and
  Tramer]{radiya2022data}
Evani Radiya-Dixit, Sanghyun Hong, Nicholas Carlini, and Florian Tramer.
\newblock Data poisoning won{\textquoteright}t save you from facial
  recognition.
\newblock In \emph{International Conference on Learning Representations}, 2022.

\bibitem[Redmon and Farhadi(2018)]{redmon2018yolov3}
Joseph Redmon and Ali Farhadi.
\newblock Yolov3: An incremental improvement.
\newblock \emph{arXiv preprint arXiv:1804.02767}, 2018.

\bibitem[Rekognition(2022)]{amazon}
Amazon Rekognition.
\newblock Comparefaces.
\newblock \url{https://tinyurl.com/j8mrvfad}, 2022.

\bibitem[Rennie et~al.(2017)Rennie, Marcheret, Mroueh, Ross, and
  Goel]{rennie2017self}
Steven~J Rennie, Etienne Marcheret, Youssef Mroueh, Jerret Ross, and Vaibhava
  Goel.
\newblock Self-critical sequence training for image captioning.
\newblock In \emph{Proceedings of the IEEE conference on computer vision and
  pattern recognition}, 2017.

\bibitem[Rhee and Lee(2013)]{Rhee2013CartoonlikeAG}
C.-H Rhee and C.H. Lee.
\newblock Cartoon-like avatar generation using facial component matching.
\newblock \emph{International Journal of Multimedia and Ubiquitous
  Engineering}, 2013.

\bibitem[Rockchip Electronics~Co.(2020)]{rv1126}
Ltd. Rockchip Electronics~Co.
\newblock \emph{Rockchip Developement Guide ISP20}, 2020.

\bibitem[Ronneberger et~al.(2015)Ronneberger, Fischer, and
  Brox]{ronneberger2015u}
Olaf Ronneberger, Philipp Fischer, and Thomas Brox.
\newblock U-net: Convolutional networks for biomedical image segmentation.
\newblock In \emph{Medical Image Computing and Computer-Assisted
  Intervention--MICCAI 2015: 18th International Conference, Munich, Germany,
  October 5-9, 2015, Proceedings, Part III 18}, 2015.

\bibitem[Ryoo et~al.(2017)Ryoo, Rothrock, Fleming, and Yang]{ryoo2017privacy}
Michael Ryoo, Brandon Rothrock, Charles Fleming, and Hyun~Jong Yang.
\newblock Privacy-preserving human activity recognition from extreme low
  resolution.
\newblock In \emph{Proceedings of the AAAI conference on artificial
  intelligence}, 2017.

\bibitem[Schroff et~al.(2015)Schroff, Kalenichenko, and
  Philbin]{schroff2015facenet}
Florian Schroff, Dmitry Kalenichenko, and James Philbin.
\newblock Facenet: A unified embedding for face recognition and clustering.
\newblock In \emph{Proceedings of the IEEE conference on computer vision and
  pattern recognition}, 2015.

\bibitem[Shan et~al.(2020)Shan, Wenger, Zhang, Li, Zheng, and
  Zhao]{shan2020fawkes}
Shawn Shan, Emily Wenger, Jiayun Zhang, Huiying Li, Haitao Zheng, and Ben~Y
  Zhao.
\newblock Fawkes: Protecting privacy against unauthorized deep learning models.
\newblock In \emph{29th USENIX security symposium (USENIX Security 20)}, 2020.

\bibitem[Shwayder(2020)]{shwayder2020clearview}
Maya Shwayder.
\newblock Clearview ai’s facial-recognition app is a nightmare for stalking
  victims.
\newblock \emph{Digital Trends}, 2020.

\bibitem[Sun et~al.(2019)Sun, Xiao, Liu, and Wang]{sun2019deep}
Ke~Sun, Bin Xiao, Dong Liu, and Jingdong Wang.
\newblock Deep high-resolution representation learning for human pose
  estimation.
\newblock In \emph{Proceedings of the IEEE/CVF conference on computer vision
  and pattern recognition}, 2019.

\bibitem[Thomos et~al.(2005)Thomos, Boulgouris, and
  Strintzis]{thomos2005optimized}
Nikolaos Thomos, Nikolaos~V Boulgouris, and Michael~G Strintzis.
\newblock Optimized transmission of jpeg2000 streams over wireless channels.
\newblock \emph{IEEE Transactions on image processing}, 2005.

\bibitem[Toshev and Szegedy(2014)]{toshev2014deeppose}
Alexander Toshev and Christian Szegedy.
\newblock Deeppose: Human pose estimation via deep neural networks.
\newblock In \emph{Proceedings of the IEEE conference on computer vision and
  pattern recognition}, 2014.

\bibitem[Tseng et~al.(2019)Tseng, Yu, Yang, Mannan, Arnaud, Nowrouzezahrai,
  Lalonde, and Heide]{Tseng2019Hyper}
Ethan Tseng, Felix Yu, Yuting Yang, Fahim Mannan, Karl~ST. Arnaud, Derek
  Nowrouzezahrai, Jean-François Lalonde, and Felix Heide.
\newblock Hyperparameter optimization in black-box image processing using
  differentiable proxies.
\newblock \emph{ACM Transactions on Graphics}, 2019.

\bibitem[ultralytics(2022)]{yolov5}
ultralytics.
\newblock yolov5.
\newblock \url{https://github.com/ultralytics/yolov5}, 2022.

\bibitem[Van~der Maaten and Hinton(2008)]{van2008visualizing}
Laurens Van~der Maaten and Geoffrey Hinton.
\newblock Visualizing data using t-sne.
\newblock \emph{Journal of machine learning research}, 9\penalty0 (11), 2008.

\bibitem[Vedantam et~al.(2014)Vedantam, Zitnick, and
  Parikh]{Vedantam2014CIDErCI}
Ramakrishna Vedantam, C.~Lawrence Zitnick, and Devi Parikh.
\newblock Cider: Consensus-based image description evaluation.
\newblock \emph{2015 IEEE Conference on Computer Vision and Pattern Recognition
  (CVPR)}, 2014.

\bibitem[Wang et~al.(2019)Wang, Qiu, Peng, Fu, and Zhu]{Wang2019AICD}
Jianbo Wang, Kai Qiu, Houwen Peng, Jianlong Fu, and Jianke Zhu.
\newblock Ai coach: Deep human pose estimation and analysis for personalized
  athletic training assistance.
\newblock In \emph{27th ACM International Conference on Multimedia}, 2019.

\bibitem[Wang et~al.(2021)Wang, Li, Zhang, and Shan]{wang2021towards}
Xintao Wang, Yu~Li, Honglun Zhang, and Ying Shan.
\newblock Towards real-world blind face restoration with generative facial
  prior.
\newblock In \emph{Proceedings of the IEEE/CVF Conference on Computer Vision
  and Pattern Recognition}, 2021.

\bibitem[Wang et~al.(2022{\natexlab{a}})Wang, Cun, Bao, Zhou, Liu, and
  Li]{wang2022uformer}
Zhendong Wang, Xiaodong Cun, Jianmin Bao, Wengang Zhou, Jianzhuang Liu, and
  Houqiang Li.
\newblock Uformer: A general u-shaped transformer for image restoration.
\newblock In \emph{Proceedings of the IEEE/CVF Conference on Computer Vision
  and Pattern Recognition}, 2022{\natexlab{a}}.

\bibitem[Wang et~al.(2022{\natexlab{b}})Wang, Yan, Yan, Chen, and
  Yang]{Wang2022CamShieldSS}
Zhiwei Wang, Yihui Yan, Yueli Yan, Huangxun Chen, and Zhice Yang.
\newblock Camshield: Securing smart cameras through physical replication and
  isolation.
\newblock In \emph{USENIX Security Symposium}, 2022{\natexlab{b}}.

\bibitem[Wang et~al.(2003)Wang, Simoncelli, and Bovik]{wang2003multiscale}
Zhou Wang, Eero~P Simoncelli, and Alan~C Bovik.
\newblock Multiscale structural similarity for image quality assessment.
\newblock In \emph{The Thrity-Seventh Asilomar Conference on Signals, Systems
  \& Computers, 2003}, 2003.

\bibitem[Wang et~al.(2004)Wang, Bovik, Sheikh, and Simoncelli]{wang2004image}
Zhou Wang, Alan~C Bovik, Hamid~R Sheikh, and Eero~P Simoncelli.
\newblock Image quality assessment: from error visibility to structural
  similarity.
\newblock \emph{IEEE transactions on image processing}, 2004.

\bibitem[Wenger et~al.(2022)Wenger, Shan, Zheng, and Zhao]{wenger2022sok}
Emily Wenger, Shawn Shan, Haitao Zheng, and Ben~Y Zhao.
\newblock Sok: Anti-facial recognition technology.
\newblock In \emph{2023 IEEE Symposium on Security and Privacy (SP)}, 2022.

\bibitem[Wu et~al.(2018)Wu, Wang, Wang, and Jin]{Wu2018TowardsPV}
Zhenyu Wu, Zhangyang Wang, Zhaowen Wang, and Hailin Jin.
\newblock Towards privacy-preserving visual recognition via adversarial
  training: A pilot study.
\newblock \emph{ArXiv}, abs/1807.08379, 2018.

\bibitem[Wu et~al.(2020)Wu, Wang, Wang, Jin, and
  Wang]{Wu2020PrivacyPreservingDA}
Zhenyu Wu, Haotao Wang, Zhaowen Wang, Hailin Jin, and Zhangyang Wang.
\newblock Privacy-preserving deep action recognition: An adversarial learning
  framework and a new dataset.
\newblock \emph{IEEE Transactions on Pattern Analysis and Machine
  Intelligence}, 2020.

\bibitem[XILINX(2022)]{gammaLUT}
AMD XILINX.
\newblock Gamma lut.
\newblock \url{https://tinyurl.com/4n4snxrv}, 2022.

\bibitem[Yang et~al.(2022)Yang, Yau, Fei-Fei, Deng, and
  Russakovsky]{Yang2022ASO}
Kaiyu Yang, Jacqueline Yau, Li~Fei-Fei, Jia Deng, and Olga Russakovsky.
\newblock A study of face obfuscation in imagenet.
\newblock In \emph{International Conference on Machine Learning}, 2022.

\bibitem[Yang et~al.(2021)Yang, Dong, Pang, Su, Zhu, Chen, and
  Xue]{Yang2021TowardsFE}
Xiao Yang, Yinpeng Dong, Tianyu Pang, Hang Su, Jun Zhu, Yuefeng Chen, and
  Hui~Wen Xue.
\newblock Towards face encryption by generating adversarial identity masks.
\newblock In \emph{2021 IEEE/CVF International Conference on Computer Vision
  (ICCV)}, 2021.

\bibitem[Yin et~al.(2021)Yin, Wang, Yao, Guo, Kong, Ding, Li, and
  Liu]{yin2021adv}
Bangjie Yin, Wenxuan Wang, Taiping Yao, Junfeng Guo, Zelun Kong, Shouhong Ding,
  Jilin Li, and Cong Liu.
\newblock Adv-makeup: A new imperceptible and transferable attack on face
  recognition.
\newblock \emph{arXiv preprint arXiv:2105.03162}, 2021.

\bibitem[Zhang et~al.(2016)Zhang, Zhang, Li, and Qiao]{zhang2016joint}
Kaipeng Zhang, Zhanpeng Zhang, Zhifeng Li, and Yu~Qiao.
\newblock Joint face detection and alignment using multitask cascaded
  convolutional networks.
\newblock \emph{IEEE signal processing letters}, 2016.

\end{thebibliography}

\newpage
\appendix

\subsection{Implementation of Baseline Approaches}
\label{app:baseline}

We implement two existing hardware-level approaches of privacy protection, i.e., using a low-resolution camera~\cite{ryoo2017privacy} and using a defocused camera~\cite{Pittaluga2015PrivacyPO}, as the baselines in Sec.~\ref{sec:person} and Sec.~\ref{sec:tradeoffs}. For a fair comparison with \ours, we finetune a specialized person detection model for each baseline approach to evaluate the performance on the protected images. In the following, we present the details of our implemented baseline approaches.

\noindent \textbf{Low Resolution.} We simulate the outputs of a low-resolution camera via image downsampling~\cite{ryoo2017privacy}, where the downsampling factor controls the intensity of privacy protection. Since the downsampled images have an invalid shape for recognition, we upsample the images to their original shape with a bilinear interpolation. 

\noindent \textbf{Defocused Blur.} We simulate the blurry images output by a defocused camera via applying Gaussian blur to the raw images~\cite{Pittaluga2015PrivacyPO}, where the kernel size and standard deviation of Gaussian blur affect the intensity of privacy protection. A larger kernel or higher standard deviation makes the Gaussian blur stronger.

\noindent \textbf{Parameter Selection.} Both the low-resolution and defocused camera have parameters that control the intensity of privacy protection. A reasonable selection of parameters is important to conduct comparisons between methods. The details of parameter selection are as follows.

When comparing the utility of person detection in Sec.~\ref{sec:person}, we select appropriate parameters to achieve similar effects of privacy protection as \ours. Specifically, the downsampling factor of the low-resolution camera is selected as 16, and the Gaussian kernel size and standard deviation of the defocused camera are 25 and 7, respectively. We use the face identification accuracy of Ada18 on CelebA to indicate the effects of privacy protection. The low-resolution and defocused camera lower the accuracy to 0.5\% and 0.6\%, respectively, which is close to \ours, i.e., 0.4\%. 

When comparing the privacy-utility trade-offs in Sec.~\ref{sec:tradeoffs}, we explore another two parameters that lower the intensity of protection to promote the utility of the targeted vision application, in addition to the one investigated in Sec.~\ref{sec:person} for each baseline method. Specifically, we set the downsampling factor of the low-resolution camera as 4/8/16, and set the Gaussian kernel size and standard deviation of the defocused camera as 9/13/15 and 3/5/7, respectively.

\newpage
\subsection{\ours Real-world Deployment}
\label{app:realworld}

The color correction matrix (CCM) is related to the light condition; hence, many ISPs such as RV1126~\cite{rv1126} employ a dynamic CCM related to the color temperature to achieve better image quality. In the following, we present how to deploy \ours on an ISP with a dynamic CCM.

Supposed that there are $n$ calibrated CCMs under the standard light conditions whose color temperatures $T$ are fixed. The tuples of calibrated CCM and color temperature are denoted as: 
$$(T_1, {CCM}_1), (T_2, {CCM}_2), \cdots, (T_n, {CCM}_n)$$
where the color temperatures $T_1, T_2, \cdots, T_n$ are in the ascending order.

For a given color temperature $T_e$ estimated by the ISP, the dynamic CCM $CCM_{ori}$ is linearly interpolated by two calibrated CCMs ${CCM}_i, {CCM}_{i+1}$ whose color temperatures $T_i, T_{i+1}$ are the nearest to the estimated one.
$$ CCM_{ori} = \frac{T_{e} - T_{i}}{T_{i+1} - T_{i}} \cdot {CCM}_{i+1}  + \frac{T_{i+1} - T_{e}}{T_{i+1} - T_{i}} \cdot {CCM}_{i}$$

To deploy the optimized CCM ${CCM}_{opt}$, we should let $CCM_{ori} \rightarrow CCM_{ori} \cdot {CCM}_{opt}$, according to the virtual imaging pipeline as presented in Sec.~\ref{sec:virtual}. Since the dynamic CCM is a linear combination of calibrated CCMs, we modify the calibrated CCMs as follows:
$$
    \left\{
        \begin{aligned}
            {CCM}_1 &\rightarrow CCM_{1} \cdot {CCM}_{opt} \\
            {CCM}_2 &\rightarrow CCM_{2} \cdot {CCM}_{opt} \\
            & \cdots \\
            {CCM}_n &\rightarrow CCM_{n} \cdot {CCM}_{opt} \\
        \end{aligned}
    \right.
$$
to achieve the deployment on a dynamic CCM.

\newpage
\subsection{Algorithm, Figures and Tables}

\begin{algorithm}[hb]
    \caption{\ours Adversarial Learning}
    \label{alg:main}
    \KwIn{Camera imaging pipeline $C$ and its ISP parameters $\theta_C$. Facial identification model $F$ and its parameters $\theta_F$. Person detection model $P$ and its parameters $\theta_P$. Face dataset $\mathcal{D}_F$. Person detection dataset $\mathcal{D}_P$.}
    \KwOut{Camera ISP parameters $\theta_C^*$. Person detection model parameters $\theta_P^*$.}
    \BlankLine
    Initialize the proxy head $H$ and its parameters $\theta_H$\;
    \Repeat{Accuracy is higher than threshold.}
    {
        \For{$i = 1,2,\dots,\text{Length of }\mathcal{D}_F$}{
            Sample data $x$ and label $y$ from $\mathcal{D}_F$\;
            $p = \textrm{Softmax}(H(F(x)))$\;
            $L_{ce} = - \log p_y$\;
            $\theta_H \leftarrow \theta_H - \alpha_H \nabla_{\theta_H} L_{ce}$\;
            $\theta_F \leftarrow \theta_F - \alpha_F \nabla_{\theta_F} L_{ce}$\;
        }
    }
    \For{$i = 1,2,\dots,maxiters$}{
        \For{$j = 1,2,\dots,m$}{
            Sample data $x$ and label $y$ from $\mathcal{D}_F$\;
            $\tilde{x} = C(x)$\;
            $p = \textrm{Softmax}(H(F(\tilde{x})))$\;
            $L_{ce} = - \log p_y$\;
            $\theta_H \leftarrow \theta_H - \alpha_H \nabla_{\theta_H} L_{ce}$\;
            $\theta_F \leftarrow \theta_F - \alpha_F \nabla_{\theta_F} L_{ce}$\;
        }
        \For{$j = 1,2,\dots,n$}{
            Sample data $x_1$ and label $y_1$ from $\mathcal{D}_F$\;
            $\tilde{x}_1 = C(x_1)$\;
            $p = \textrm{Softmax}(H(F(\tilde{x}_1)))$\;
            $L_{ns} = - \log (1 - p_y)$\;
            Sample data $x_2$ and label $y_2$ from $\mathcal{D}_P$\;
            $\tilde{x}_2 = C(x_2)$\;
            $cls, box = P(\tilde{x}_2)$\;
            $L_{det} = L_{cls}(cls, y_2) + L_{box}(box, y_2)$\;
            $\theta_C \leftarrow \theta_C - \alpha_C \nabla_{\theta_C} (L_{ns} + \omega L_{det})$\;
            $\theta_P \leftarrow \theta_P - \alpha_P \nabla_{\theta_P} L_{det}$\;
        }
    }
\end{algorithm}

\begin{figure}[h]
    \centering
    \subfigure[\textit{``Two women playing a video game in a living room.''}]{
        \includegraphics[width=.3\linewidth,height=0.11\textwidth]{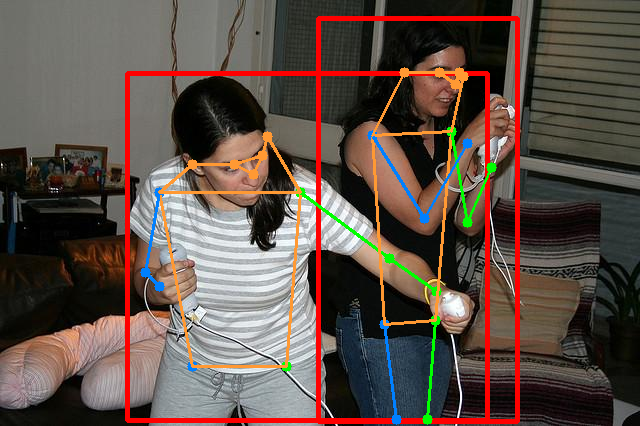}}
    \hfill
    \subfigure[\textit{``A man and a woman sitting at a table with food.''}]{
        \includegraphics[width=.3\linewidth,height=0.11\textwidth]{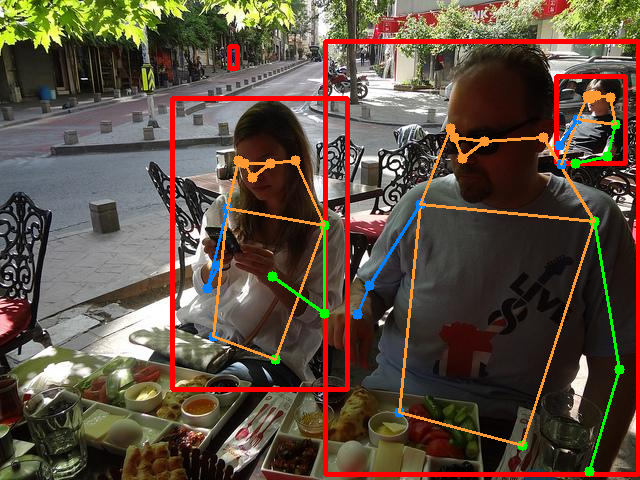}}
    \hfill
    \subfigure[\textit{``A group of men standing next to each other.''}]{
        \includegraphics[width=.3\linewidth,height=0.11\textwidth]{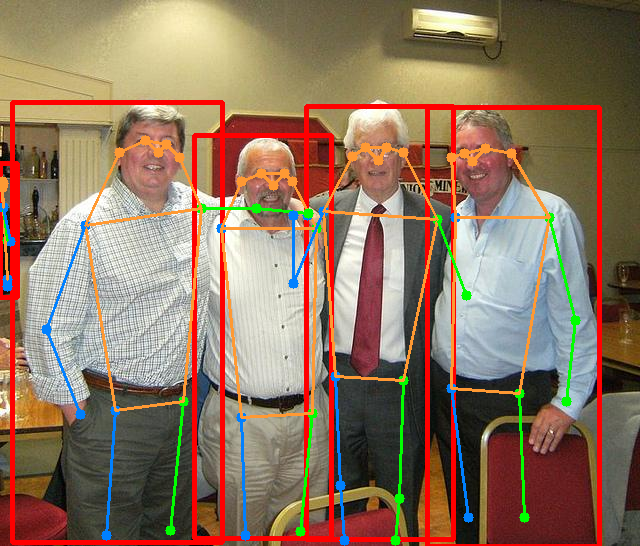}}

    \subfigure[\textit{``Two people playing a video game in a living room.''}]{
        \includegraphics[width=.3\linewidth,height=0.11\textwidth]{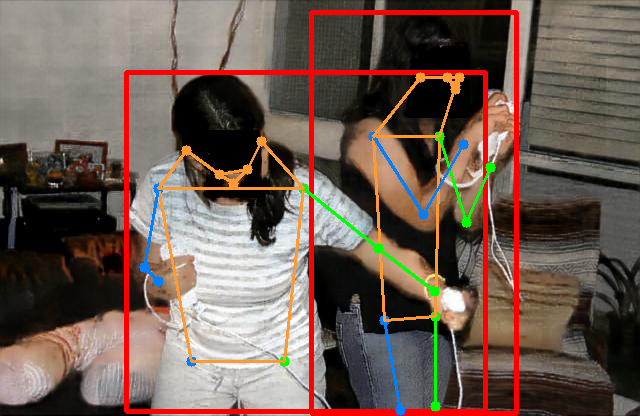}}
    \hfill
    \subfigure[\textit{``A man and a woman sitting at a table with food.''}]{
        \includegraphics[width=.3\linewidth,height=0.11\textwidth]{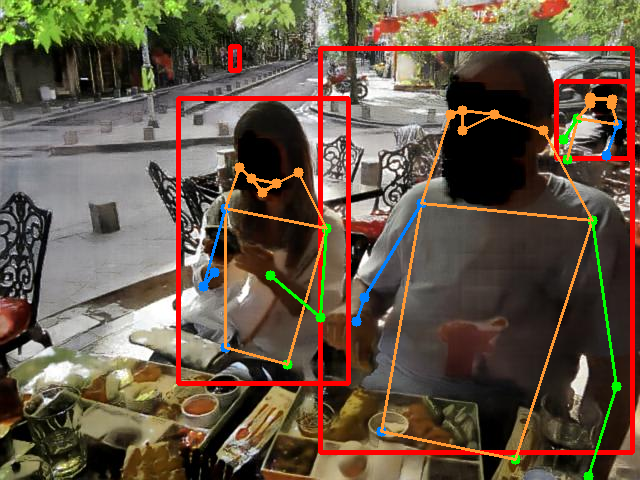}}
    \hfill
    \subfigure[\textit{``A group of men standing next to each other.''}]{
        \includegraphics[width=.3\linewidth,height=0.11\textwidth]{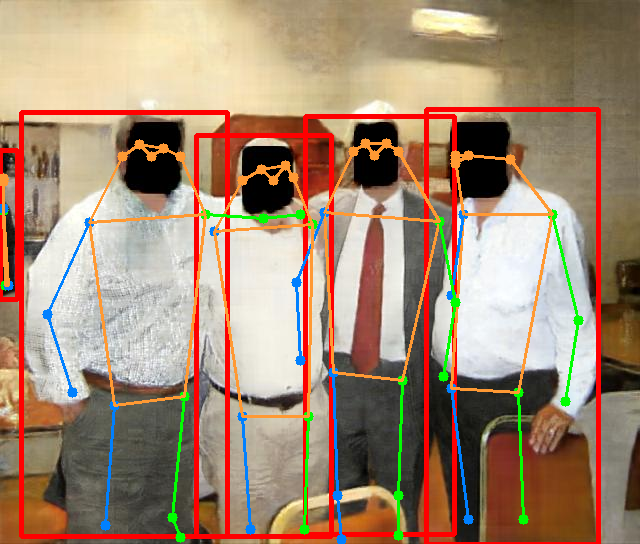}}
    \caption{Qualitative results of three HAR vision applications: (1) person detection (plotted with red boxes), (2) human pose estimation (plotted with points and skeletons), and (3) image captioning (displayed with short captions below pictures). Fig.~(a)-(c) show the recognition results of the raw images, and Fig.~(d)-(f) show the results of the enhanced images of \ours. Related quantitative results and analyses can be found in Sec.~\ref{sec:utilityGeneral}.}
    \label{fig:coco}
\end{figure}


\begin{figure}[h]
    \centering
    \subfigure[simulated captured image]{
        \includegraphics[width=.47\linewidth,height=0.13\textwidth]{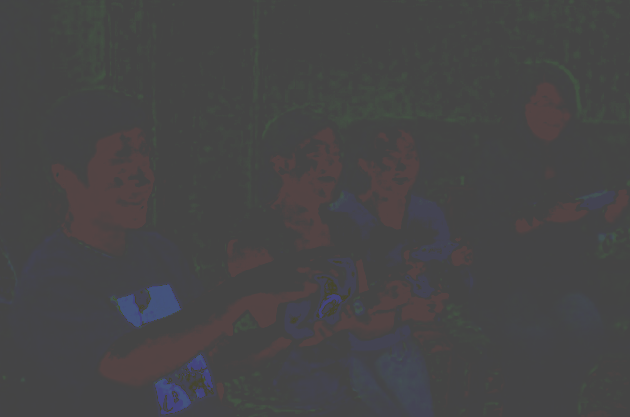}}
    \hfill
    \subfigure[real-world captured image]{
        \includegraphics[width=.47\linewidth,height=0.13\textwidth]{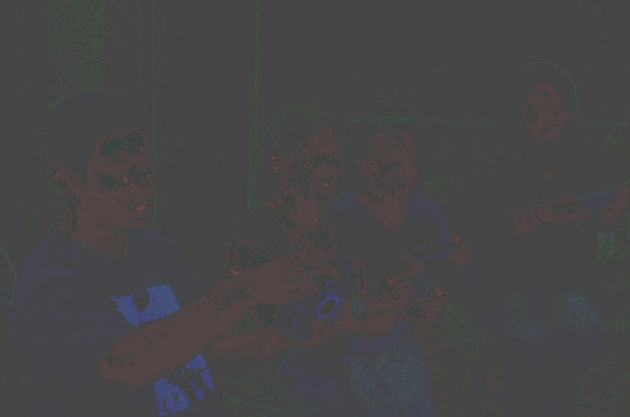}}
    \caption[short]{Captured image in simulation and real-world. Related quantitative results can be found in Sec.~\ref{sec:realworldimage}.}
    \label{fig:rw1}
\end{figure}

\begin{figure}[h]
    \centering
    \subfigure[raw image]{
        \includegraphics[width=.47\linewidth,height=0.13\textwidth]{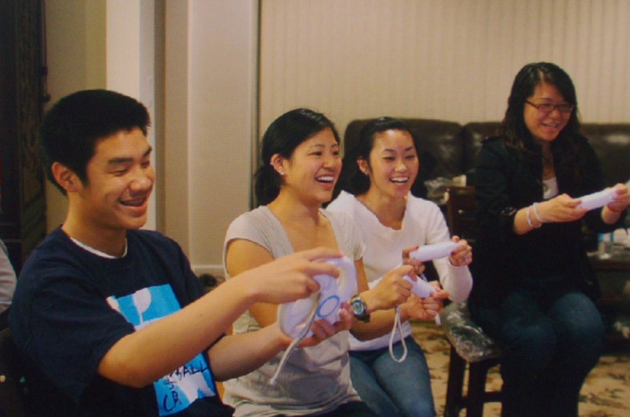}}
    \hfill
    \subfigure[\ours enhanced image]{
        \includegraphics[width=.47\linewidth,height=0.13\textwidth]{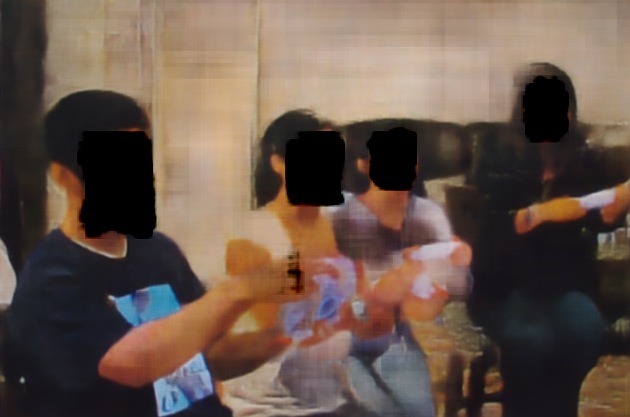}}
    \caption[short]{Real-world raw image and \ours enhanced image. Related quantitative results can be found in Sec.~\ref{sec:realworldimage}.}
    \label{fig:rw2}
\end{figure}

\begin{table}[h]
    \centering
    \caption{Specifications of tested Android smartphones}
    \label{tab:phone}
    \begin{tabular}{@{\extracolsep{6pt}}cccc@{}}
    \toprule
    Device Model & OS & Android version & \ours \\ \midrule
    Google Pixel & stock Android & 10 & \cmark \\
    Samsung S20 FE & One UI 3.1 & 11 & \cmark \\
    Huawei Nova 4 & EMUI 10.0.0 & 10 & \cmark \\ 
    OPPO Find X5 Pro & ColorOS 13.1 & 13 & \cmark \\
    iQOO Neo5 SE & OriginOS 3 & 13 & \cmark \\ 
    iQOO Neo6 SE & OriginOS 3 & 13 & \cmark \\ 
    Redmi K30S Ultra & MIUI 14.0.5 & 12 & \cmark \\
    MEIZU 16th Plus & Flyme 8.1.8.0A & 8 & \cmark \\
    \bottomrule
    \end{tabular}
\end{table}

\clearpage

\appendices
\section{Artifact Appendix}

\subsection{Description \& Requirements}

This artifact includes the major evaluation experiments presented in \ours paper. To largely shorten the required time for evaluation, we omit the data preprocessing and training procedure of \ours; instead, we provide the links to processed datasets, optimized ISP parameters and vision application models in our instruction \textsl{README.md}.

\subsubsection{How to access}
We uploaded the artifact to Zenodo, which is a platform that offers permanent storage with a DOI. The DOI link is \url{https://doi.org/10.5281/zenodo.10156141}. 

\subsubsection{Hardware dependencies}
Our experiments can be run on most commodity desktop/laptop machines. However, since there are several neural network models used for the evaluation, it is necessary to run the experiments on a commodity GPU (e.g., NVIDIA RTX 3090) to shorten the whole progress. The recommended GPU memory is more than 12 GB. To solve the memory limitation problem, it is feasible to lower the batch size of the data loader. To storage the large image datasets, it is recommended to spare more than 64 GB disk space.

\subsubsection{Software dependencies} 
\begin{itemize}
    \item A recent Linux OS (e.g., Ubuntu 18.04/20.04/22.04)
    \item CUDA driver version higher than 11.3
    \item Python 3.9 (Recommended to install Anaconda)
    \item Other python packages listed in \textsl{requirements.txt}
\end{itemize}

\subsubsection{Benchmarks} 
\begin{itemize}
    \item CelebA face dataset, where the faces have been cropped and aligned by a face detector MTCNN.
    \item LFW face dataset, where the faces have been cropped and aligned by a face detector MTCNN.
    \item MS COCO detection 2017 dataset.
\end{itemize}

\subsection{Artifact Installation \& Configuration}

Please follow the instructions of \textsl{Python Environment Setup} and \textsl{Data Preparation} contained in \textsl{README.md}. We also provide a bash script \textsl{start.sh} to automate the procedures of data preparation and python environment setup. Please follow the instructions of \textsl{Semi-Auto Setup} contained in \textsl{README.md}.

\subsection{Experiment Workflow}

All the experiments are invoked by executing the corresponding python script under the folder \textsl{src/evaluation/}. The experimental results are organized and saved to the folder \textsl{results/} with a CSV format except for E5. In E5, the results (e.g., the best attack success rate) are printed in the console.

\subsection{Major Claims}

\ours has the following major claims:

\begin{itemize}
    \item (C1): \ours can largely lower the face identification accuracies on both the captured images and the enhanced images [Sec.~VI-B1]. \ours can be effective across various facial recognition models [Sec.~VI-B2], various face identification classifiers [Sec.~VI-B3], and various face datasets [Sec.~VI-B4]. This is proven by the experiment (E1) whose results are illustrated in [Table~I].
    \item (C2): \ours adversarial learning scheme is important to generalize on those unseen/black-box facial recognition models [Sec.~VI-D]. If the attacker update step is removed, the protection may be ineffective on various unseen models. This is proven by the experiment (E2) whose results are shown in [Table~II].
    \item (C3): \ours can maintain the utility of the target vision application, i.e., person detection, on the captured images [Sec.~VI-E1]. \ours performs better than the baselines, i.e., the low-resolution and defocused approach, as for various metrics of person detection. It is proven by the experiment (E3) whose results are presented in [Table~III].
    \item (C4): \ours image enhancer can largely improve the image quality to make it friendly for possible human viewers to recognize the happened human activities [Sec.~VI-E2]. It is proven by the experiment (E4) whose results are presented in [Table~IV].
    \item (C5): \ours are resistant to the white-box adaptive attacks [Sec.~VII-C]. Even if the adversary can re-train the facial recognition model with the full knowledge of \ours including the exact deployed parameters, \ours can still reduce the face identification accuracies to a low level. This is proven by the experiment (E5) whose results are shown in [Table~VII].
\end{itemize}

\subsection{Evaluation}

\subsubsection{Anti-Facial Recognition Experiment (E1)}
[15 human-minutes + 5 compute-hours]: 
In this experiment, we evaluate the performance of anti-facial recognition (AFR) on the protected images, i.e., the captured images and the enhanced images. A lower face identification accuracy indicates a better AFR effect.

\textit{[How to]}
The experiments can be conducted by running a python file named \textsl{exp1.py}.

\textit{[Preparation]}
None.

\textit{[Execution]}
First, activate your python environment if you use Anaconda. Second, change your current directory to \textsl{src/}. Third, run the python script \textsl{evaluation/exp1.py}. Finally, check the saved result in \textsl{results/1.csv}. For reference, the detailed commands are listed in \textsl{README.md}.

\textit{[Results]}
The results will be saved to a CSV file whose path is \textsl{results/1.csv}. The format is the same as [Table~I]. Since there are randomness in our evaluation protocol, as discussed in the \textbf{evaluation protocol} part in [Sec.~VI-A], we expect that there are tiny numerical differences ($<0.01$) between the reproduced results and the reported ones in this paper.

\subsubsection{Ablation Study Experiment (E2)}
[15 human-minutes + 0.5 compute-hours]: 
In this experiment, we make an ablation study on the adversarial learning framework. We remove the Attacker Update Step in adversarial learning, and obtain a set of parameters \textsl{checkpoints/ablation.pt} against a white-box model Ada18 \textsl{checkpoints/whitebox.pt}. Here, we evaluate the ablated parameters against other nine black-box models. As reported in our paper, the ablated parameters would perform much worse on those black-box models than on the white-box models.

\textit{[How to]}
The experiments can be conducted by running a python file named \textsl{exp2.py}.

\textit{[Preparation]}
None.

\textit{[Execution]}
First, activate your python environment if you use Anaconda. Second, change your current directory to \textsl{src/}. Third, run the python script \textsl{evaluation/exp2.py}. Finally, check the saved result in \textsl{results/2.csv}. For reference, the detailed commands are listed in \textsl{README.md}.

\textit{[Results]}
The results are saved to a CSV file whose path is \textsl{results/2.csv}. The format is the same as [Table~II]. The last column in [Table~II] is exactly same as the first row in [Table~I]; hence, it has been validated in (E1). Since there are randomness in our evaluation protocol, as discussed in the \textbf{evaluation protocol} part in [Sec.~VI-A], we expect that there are tiny numerical differences ($<0.01$) between the reproduced results and the reported ones in this paper. Note that the Accuracy ratio (Black-box/White-box) may have a bit larger difference ($<1.0$) because its denominator is quite small.

\subsubsection{Vision Application Performance Experiment (E3)}
[15 human-minutes + 0.5 compute-hours]: 
In this experiment, we evaluate the target vision application performances with \ours and with two baseline methods. To have a fair comparison, we finetune the person detector model \textsl{yolov5m} for each method. The finetuned model weights are saved as \textsl{captured.pt}, \textsl{lowres.pt}, and \textsl{defocus.pt}, respectively. We calculate common metrics of object detection, e.g., AP, Precision, Recall, F1, etc., for each method.

\textit{[How to]}
The experiments can be conducted by running a python file named \textsl{exp3.py}.

\textit{[Preparation]}
None.

\textit{[Execution]}
First, activate your python environment if you use Anaconda. Second, change your current directory to \textsl{src/}. Third, run the python script \textsl{evaluation/exp3.py}. Finally, check the saved result in \textsl{results/3.csv}. For reference, the detailed commands are listed in \textsl{README.md}.

\textit{[Results]}
The results are saved to a CSV file whose path is \textsl{results/3.csv}. The format is the same as [Table~III]. Due to the randomness of the data loader, we expect that there are trivial numerical differences between the reproduced results and the reported ones in this paper.

\subsubsection{Image Quality Assessment Experiment (E4)}
[15 human-minutes + 0.2 compute-hours]: 
In this experiment, we evaluate the image quality of both the captured images and the enhanced images. The metrics include RMSE, PSNR, SSIM, and MS-SSIM. The enhanced images have a much higher quality than the captured images.

\textit{[How to]}
The experiments can be conducted by running a python file named \textsl{exp4.py}.

\textit{[Preparation]}
None.

\textit{[Execution]}
First, activate your python environment if you use Anaconda. Second, change your current directory to \textsl{src/}. Third, run the python script \textsl{evaluation/exp4.py}. Finally, check the saved result in \textsl{results/4.csv}. For reference, the detailed commands are listed in \textsl{README.md}.

\textit{[Results]}
The results are saved to a CSV file whose path is \textsl{results/4.csv}. The format is the same as [Table~IV]. Due to the randomness of the data loader, we expect that there are trivial numerical differences between the reproduced results and the reported ones in this paper.

\subsubsection{White-box Adaptive Attack Experiment (E5)}
[15 human-minutes + 1.5 compute-hours per run]: 
In this experiment, we evaluate the AFR performance of \ours against a white-box adversary who re-trains the facial recognition model. Since the re-training of 40 models requires a lot of time (about 60 compute-hours), we scale down the experiment to a single model.

\textit{[How to]}
The experiments can be conducted by running a python file named \textsl{exp5.py}. The python script has the following arguments: 
\begin{itemize}
    \item \texttt{--backbone n}: \texttt{n} is an integer from 0 to 9, where each integer represents a different facial recognition model respectively.
    \item \texttt{--head n}: \texttt{n} is an integer from 0 to 1, where 0 stands for Softmax and 1 stands for ArcFace. 
    \item \texttt{--mode n}: \texttt{n} is an integer from 0 to 1, where 0 stands for finetuning and 1 stands for training-from-scratch. 
\end{itemize}

\textit{[Preparation]}
None.

\textit{[Execution]}
First, activate your python environment if you use Anaconda. Second, change your current directory to \textsl{src/}. Third, run the python script \textsl{evaluation/exp5.py}. Finally, check the printed result in the console. For reference, the detailed commands are listed in \textsl{README.md}.

\textit{[Results]}
The results will be printed in your console. The printed \textsl{Highest Accuracy} is the number presented in [Table~VII]. As discussed in [Sec.~VII-C1], the model accuracy may be very instable. When the model goes overfitting, the accuracy may approach to 0\%. We expect that the highest accuracy is less than 20\% in most trials.

\end{document}